\documentclass{article}

\PassOptionsToPackage{numbers}{natbib}
\usepackage[preprint]{neurips_2026} 

\usepackage[utf8]{inputenc} 
\usepackage[T1]{fontenc}    
\usepackage{hyperref}       
\usepackage{url}            
\usepackage{booktabs}       
\usepackage{amsfonts}       
\usepackage{nicefrac}       
\usepackage{microtype}      
\usepackage{xcolor}         
\usepackage{graphicx}

\usepackage{amsmath}
\usepackage{multirow}
\usepackage{pifont}
\usepackage{subcaption}
\usepackage[table,xcdraw]{xcolor}
\usepackage{wrapfig}
\usepackage{float}
\usepackage{placeins}
\usepackage{subcaption}

\title{Align3D-AD: Cross-Modal Feature Alignment and Dual-Prompt Learning for Zero-shot 3D Anomaly Detection}

\author{
Letian Bai$^{1}$, Xuanming Cao$^{1}$, Juan Du$^{1,2}$\thanks{Corresponding author.}, Chengyu Tao$^{3}$ \\
$^{1}$ Smart Manufacturing Thrust, The Hong Kong University of Science and Technology (Guangzhou) \\
$^{2}$ The Hong Kong University of Science and Technology \\
$^{3}$ College of Mechanical and Vehicle Engineering, Hunan University \\
\texttt{lbai799@connect.hkust-gz.edu.cn, xcao743@connect.hkust-gz.edu.cn} \\
\texttt{juandu@ust.hk, taochengyu@hnu.edu.cn}
}

\begin{document}

\maketitle

\begin{abstract}
Zero-shot 3D anomaly detection aims to identify anomalies without access to training data from target categories. 
However, existing methods mainly rely on projecting 3D observations into multi-view representations that primarily capture geometric cues rather than realistic visual semantics and process them with vision encoders pretrained on RGB data, leading to a significant domain gap between the encoder and the projected representations. 
To address this issue, we propose Align3D-AD, a unified two-stage framework that leverages the RGB modality from auxiliary categories as cross-modal guidance for zero-shot 3D anomaly detection. 
First, we introduce a cross-modal feature alignment paradigm that maps rendering features into the RGB semantic space. 
Unlike prior works that implicitly rely on pretrained encoders, our method enables direct semantic transfer from RGB observations. 
A semantic consistency reweighting strategy is further introduced to refine feature alignment by reweighting local regions according to holistic semantic consistency.
Second, we propose a modality-aware prompt learning framework with dual-prompt contrastive alignment. 
By assigning independent prompts to RGB-aligned and rendering features, our method captures complementary semantics across modalities, while the contrastive alignment further enhances prompt representations to improve discriminability.
Extensive experiments on MVTec3D-AD, Eyecandies, and Real3D-AD demonstrate that Align3D-AD consistently outperforms existing zero-shot methods under both one-vs-rest and cross-dataset settings, highlighting its generalization capability and robustness.
\textbf{Code and the dataset will be made available once our paper is accepted.}
\end{abstract}

\section{Introduction}
\label{sec:introduction}

Surface anomaly detection plays a crucial role in industrial quality control \cite{tao2025pointsgrade,du20253d, G2SF, cao2023anomaly, bergmann2020uninformed, cohen2020sub}. 
Compared to traditional 2D image inspection, which is highly susceptible to lighting conditions and lacks 3D geometric details, automated detection based on 3D point clouds has garnered increasing attention \cite{cao2025iaenet}. 
Due to the extreme scarcity of anomalies in industrial scenarios, most prevailing methods fall into the category of unsupervised approaches, learning the feature distribution of normal samples primarily through memory banks \cite{roth2022towards,horwitz2023back,wang2023multimodal} or teacher-student architectures \cite{bergmann2023anomaly,rudolph2023asymmetric}. 
However, these unsupervised methods still rely heavily on a large amount of normal data from the target category. When faced with practical industrial constraints such as privacy protection or the absence of new product prototypes \cite{zhou2023anomalyclip,PointAD, gu2024anomalygpt, gu2024filo}, training data for the target category is often completely inaccessible, prompting the emergence of zero-shot 3D anomaly detection tasks. 
Specifically, the paradigm seeks to train a general model using auxiliary data, thereby obviating the requirement for training samples from the target category \cite{GS-CLIP}.

\begin{figure}[t]
    \centering
    \hspace{-0.15\linewidth}
    \begin{subfigure}[t]{0.47\linewidth}
        \centering
        \includegraphics[height=4cm]{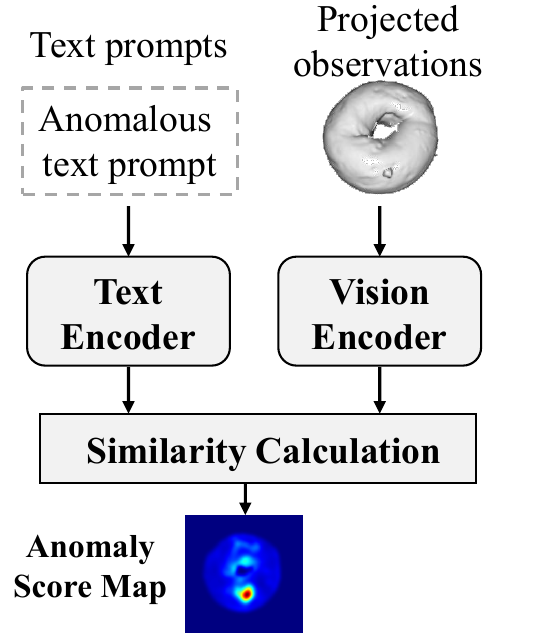}
        \caption{Zero-shot 3D anomaly detection paradigm.}
        \label{fig:motivation_a}
    \end{subfigure}
    \begin{subfigure}[t]{0.47\linewidth}
        \centering
        \includegraphics[height=4cm]{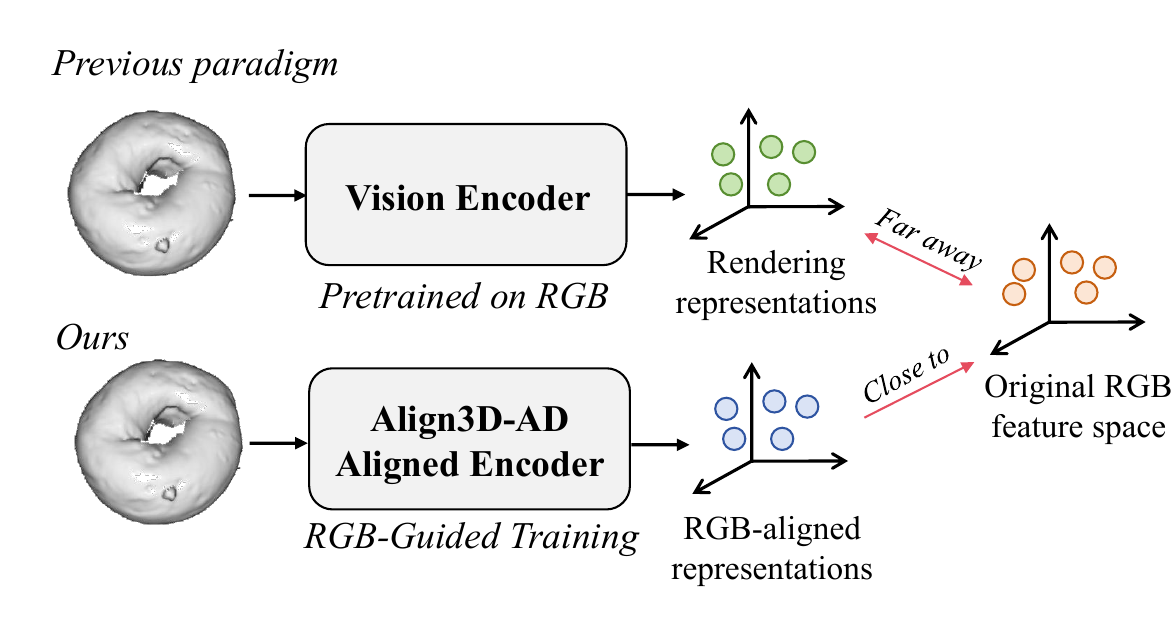}
        \caption{Implicit domain gap reduction with cross-modal alignment.}
        \label{fig:motivation_b}
    \end{subfigure}
    \caption{Motivation of Align3D-AD. (a) An overview of the existing framework for zero-shot 3D anomaly detection. (b) A comparison between the previous paradigm and the proposed method. By incorporating RGB guidance during training, the proposed model yields RGB-aligned representations, thereby effectively mitigating the domain gap inherent in the vision encoder.}
    \label{fig:motivation}
\end{figure}

Although zero-shot anomaly detection has been extensively investigated in the 2D domain \cite{jeong2023winclip,zhou2023anomalyclip,cao2024adaclip,qu2025bayesian}, research on 3D remains relatively limited, with only a few recent attempts. For example, PointAD \cite{PointAD} integrates the hybrid representation learning of points and pixels to comprehend structural anomalies, while GS-CLIP \cite{GS-CLIP} dynamically generates geometry-aware text prompts embedded with 3D priors. 
Fundamentally, this prevailing paradigm relies on the projection of 3D point clouds into multi-view 2D observations \cite{cao2024complementary}, aiming to transfer the strong generalization capability of the vision-language model (CLIP) \cite{CLIP} to the 3D domain (as illustrated in Figure \ref{fig:motivation_a}).
However, the strong semantic understanding capability of CLIP is established in the RGB feature space characterized by rich appearance and texture cues, whereas multi-view renderings generated from 3D point clouds are highly abstract and contain exclusively geometric information, thereby leading to a critical domain gap in this paradigm.
This cross-modal inconsistency makes it difficult for the model to extract deep discriminative features when provided solely with geometric inputs.

To bridge this gap, we propose Align3D-AD, a unified two-stage framework that leverages the RGB modality as cross-modal guidance during training. 
The core motivation is to utilize the RGB modality as the cross-modal guidance during the training stage (as shown in Figure \ref{fig:motivation_b}). 
Specifically, the first stage introduces the feature aligner to project rendering features into the RGB feature space, further refined by a semantic consistency reweighting strategy activated in the later phase of training.
In the second stage, we propose a modality-aware prompt learning mechanism that assigns independent prompts to both aligned and rendering features, thereby capturing complementary anomaly characteristics and semantics.
A dual-prompt contrastive alignment is further introduced to improve the consistency and discrimination capability of prompt representations.

The main contributions of this paper are summarized as follows:

\begin{itemize}
\item We propose Align3D-AD, a unified two-stage framework for zero-shot 3D anomaly detection. By introducing the RGB modality as cross-modal guidance during training, the framework learns RGB-aligned semantic representations for rendering features, enabling the model to exploit rich visual semantics even when only 3D observations are available.

\item We design a simple yet effective semantic consistency reweighting strategy. 
This mechanism adaptively assigns local alignment weights by aggregating cross-modal semantic consistency across patches, emphasizing regions with cross-modal discrepancies and enabling more robust and effective feature alignment.

\item We propose a modality-aware prompt learning scheme with dual-prompt contrastive alignment. 
By assigning independent text prompts to both RGB-aligned and rendering branches, the framework captures modality-specific semantics and further improves the effectiveness of prompt representation learning through contrastive alignment.

\item Extensive experiments on MVTec3D-AD \cite{MVTec3D-AD}, Eyecandies \cite{Eyecandies}, and Real3D-AD \cite{Real3D-AD} across diverse evaluation settings demonstrate that Align3D-AD consistently outperforms existing zero-shot methods, with only rendering observations required during inference.

\end{itemize}

The remainder of this paper is organized as follows. 
Section \ref{sec:related work} reviews the literature on unsupervised 3D anomaly detection and zero-shot anomaly detection in both 2D and 3D scenarios.
Section \ref{sec: Align3D-AD} introduces the proposed Align3D-AD framework in detail. 
Section \ref{sec:experiment} provides the experimental setup and evaluation results. 
Finally, Section \ref{sec:conclusion} concludes the paper, followed by discussions on limitations and broader impacts.

\section{Related Work}
\label{sec:related work}

\subsection{Unsupervised 3D Anomaly Detection}
Unsupervised 3D anomaly detection predominantly learns the feature distribution of normal samples through feature-based or reconstruction-based paradigms \cite{cao2025iaenet}. Feature-based methods evaluate anomalies by measuring representational deviations from established normal samples \cite{du20253d}. Within this category, memory bank approaches \cite{horwitz2023back,wang2023multimodal,cao2024complementary,chu2023shape} store nominal features extracted from 3D descriptors or multi-view renderings to compute the feature distance of test samples. Concurrently, knowledge distillation frameworks \cite{bergmann2023anomaly,rudolph2023asymmetric} utilize teacher-student architectures to quantify representational discrepancies. In contrast, reconstruction-based approaches \cite{li2024towards,zavrtanik2024cheating} identify anomalies by evaluating reconstruction errors. 
Nevertheless, all the unsupervised paradigms strictly necessitate extensive collections of normal training data, exposing a critical limitation that zero-shot methodologies aim to resolve.

\subsection{Zero-shot 2D Anomaly Detection}
To address the inaccessibility of normal samples, zero-shot anomaly detection has been extensively explored using the vision-language model (CLIP) \cite{CLIP}. WinCLIP \cite{jeong2023winclip} pioneers this by leveraging the native image-text alignment of CLIP for tuning-free anomaly detection. To overcome the limitations of hand-crafted prompts, AnomalyCLIP \cite{zhou2023anomalyclip} learns object-agnostic text templates to capture generic anomaly semantics across domains, while AdaCLIP \cite{cao2024adaclip} introduces a hybrid learnable prompt mechanism to adaptively adjust the feature space of CLIP for varying inputs. To tackle the unpredictability of anomalies, Bayes-PFL \cite{qu2025bayesian} models the posterior distribution of prompts via a Bayesian framework, significantly enhancing model robustness compared to deterministic approaches. Furthermore, AA-CLIP \cite{AA-CLIP} constructs decoupled text anchors to guide the alignment of patch-level visual features, effectively improving the inter-class discriminability between normal and anomalous representations.

\subsection{Zero-shot 3D Anomaly Detection}
Zero-shot 3D anomaly detection methods also leverage CLIP, wherein 3D data are projected and rendered into images \cite{cao2024complementary} to satisfy the input requirements of CLIP. PointAD \cite{PointAD} projects 3D point clouds into multi-view 2D renderings and integrates hybrid representation learning of points and pixels, thereby transferring the strong generalization capability of CLIP to zero-shot 3D anomaly detection. 
Subsequently, MVP-PCLIP \cite{MVP-PCLIP} converts 3D point clouds into multi-view depth maps and introduces learnable visual and adaptive text prompting to fine-tune the vision-language model, thereby enhancing detection performance. 
Furthermore, GS-CLIP \cite{GS-CLIP} dynamically generates geometry-aware text prompts embedded with 3D geometric priors and synergistically fuses multi-view representations of rendering images and depth maps, enabling the model to accurately identify complex 3D geometric anomalies. 
However, these methods overlook the domain gap inherent in the rendering modality, which significantly degrades the representational capacity of visual features. 
In contrast, the proposed Align3D-AD framework effectively bridges this domain gap and achieves superior detection results.

\begin{figure}[!t]
\centering
\includegraphics[width=\linewidth]{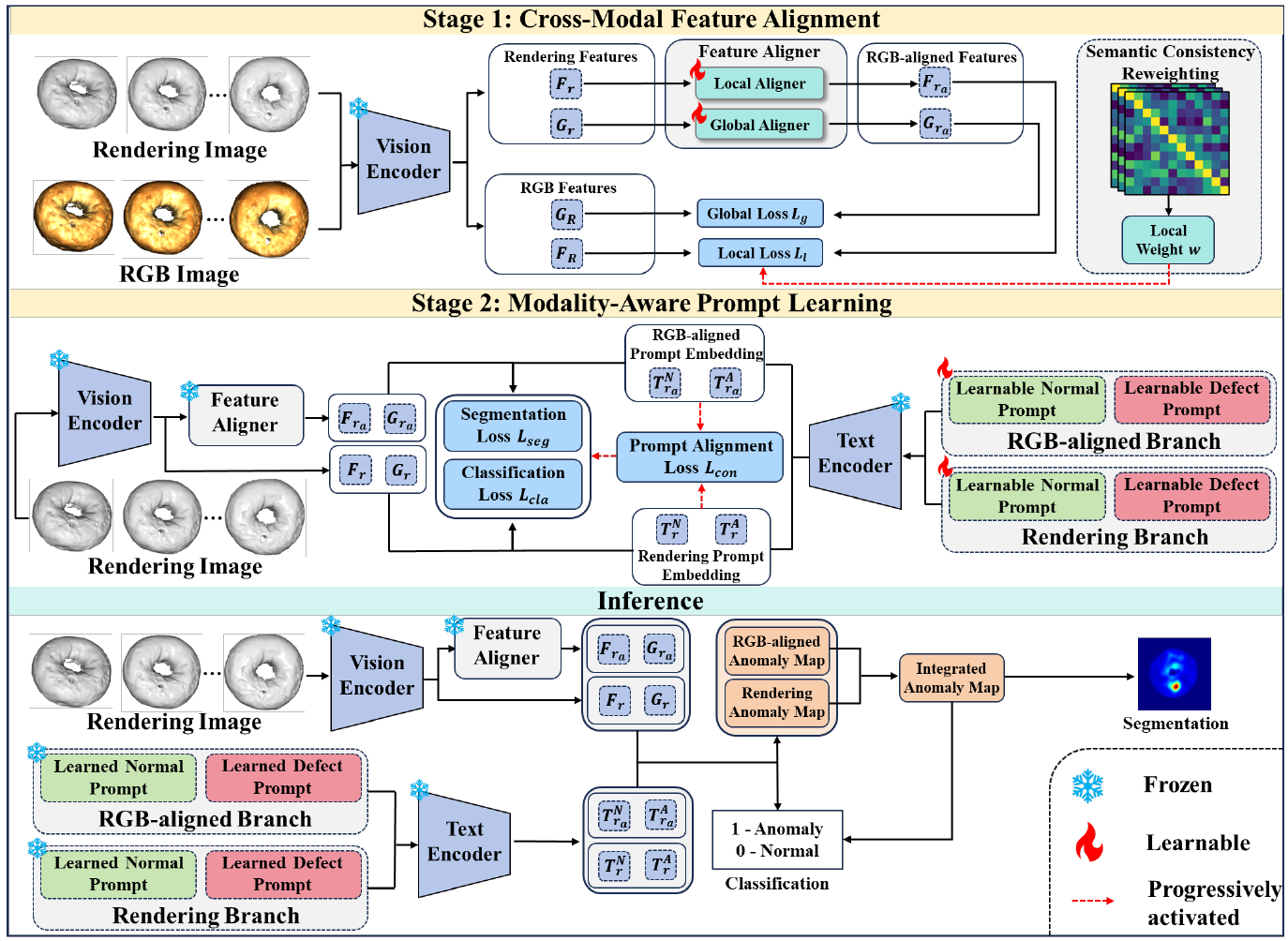}
\caption{
Overall framework of Align3D-AD. 
Our model consists of two stages: Cross-Modal Feature Alignment and Modality-Aware Prompt Learning.
Stage 1 aligns rendering and RGB features in the RGB feature space, further refined by the semantic consistency reweighting strategy. 
Stage 2 performs modality-aware prompt learning for both RGB-aligned and rendering branches, with the dual-prompt contrastive alignment activated in the later phase to enhance prompt discriminability.
During inference, anomaly predictions from both branches are integrated to produce the final anomaly detection results.
}
\label{fig:Overall Pipeline}
\end{figure}

\section{Align3D-AD}
\label{sec: Align3D-AD}

\subsection{Overview}
\label{subsec:overview}

We propose Align3D-AD, a unified two-stage framework for zero-shot 3D anomaly detection that bridges the domain gap between rendering and RGB modalities and enhances anomaly discrimination through modality-aware prompt learning.
An overview of the framework is shown in Figure \ref{fig:Overall Pipeline}.
In the first stage, we perform cross-modal feature alignment to project rendering features into the RGB feature space through the learnable feature aligner. 
The semantic consistency reweighting strategy further improves the alignment by reweighting rendering patches based on the holistic semantic consistency across modalities. 
In the second stage, we perform modality-aware prompt learning for both RGB-aligned and rendering branches. 
To further improve the representational capability of modality-aware prompts, the dual-prompt contrastive alignment is incorporated in the later phase.
By combining cross-modal alignment with structured prompt learning, Align3D-AD leverages complementary geometric and semantic cues to learn feature representations for zero-shot 3D anomaly detection.

\vspace{-6pt}

\subsection{Feature Extraction}
\label{subsec:feature extraction}

Unlike existing zero-shot 3D anomaly detection methods that rely solely on projected observations derived from 3D point clouds, we incorporate RGB images during the training stage to provide complementary semantic cues. 
Specifically, given a 3D point cloud $D_{3d}$ with $P_n$ points, we render it from $V$ viewpoints, obtaining multi-view rendering images $\{I_i^{r}\}_{i=1}^{V}$ and corresponding RGB images $\{I_i^{R}\}_{i=1}^{V}$. Here, $I_i^{m} \in \mathbb{R}^{H \times W}$ denotes the observation of modality $m \in \{R, r\}$ at the $i$-th view. 
A shared vision encoder extracts features from each view independently, producing the global feature $G_i^{m} \in \mathbb{R}^{d}$ and local features $F_i^m = [f_{i,1}^m, f_{i,2}^m, \dots, f_{i,N}^m]^{\top} \in \mathbb{R}^{N \times d}$, where $f_{i,n}^{m} \in \mathbb{R}^{d}$ corresponds to the $n$-th patch, $N$ denotes the number of patches, and $d$ is the feature dimension.

\vspace{-6pt}

\subsection{Cross-modal Feature Alignment}

\textbf{Feature alignment.}
Most large-scale vision encoders are trained on RGB images, resulting in a domain gap between rendering and RGB modalities.
To bridge this gap, we project rendering features into the RGB semantic space through the learnable feature aligner, enabling better learned semantic representations and reducing cross-modal discrepancies.

Formally, the rendering global feature is projected as $G_i^{r_a} = \mathcal{A}_g(G_i^{r})$, and the local features are projected as $F_i^{r_a} =  \mathcal{A}_l(F_i^{r})$, 
where $r_a$ denotes the RGB-aligned representation, i.e., rendering features projected into the RGB feature space, and $\mathcal{A}_g(\cdot)$ and $\mathcal{A}_l(\cdot)$ denote learnable MLPs.
To enforce cross-modal consistency, we align the projected features with their RGB counterparts using cosine similarity. 
The global alignment loss $\mathcal{L}_g$ and the local alignment loss $\mathcal{L}_l$ are defined as:
\begin{equation}
\mathcal{L}_g = \frac{1}{V} \sum_{i=1}^{V} ( 1 - \langle G_i^{r_a},G_i^{R} \rangle), \quad
\mathcal{L}_l = \frac{1}{V} \sum_{i=1}^{V} \frac{1}{N} \sum_{n=1}^{N} ( 1 - \langle f_{i,n}^{r_a},f_{i,n}^{R} \rangle),
\end{equation}
where $\langle \cdot,\cdot \rangle$ denotes cosine similarity.
The overall alignment loss is defined as $\mathcal{L}_{\text{align}} = \mathcal{L}_g + \mathcal{L}_l$, which enforces global and local feature consistency simultaneously.

\textbf{Semantic consistency reweighting.}
Although feature alignment enforces consistency in the RGB feature space, it relies on local matching and treats all patches equally, which may be suboptimal in the presence of cross-modal discrepancies.
This motivates us to move beyond local correspondence and incorporate holistic semantic information.
To this end, we propose a simple yet effective reweighting strategy that adaptively modulates the contribution of each rendering patch based on the holistic semantic consistency with the RGB modality.
To ensure stability, the reweighting strategy is applied only to the local alignment loss $\mathcal{L}_l$ and is activated in the later phase of cross-modal feature alignment.

Specifically, we compute the semantic consistency matrix $C_i = F_i^{r} (F_i^{R})^{\top} \in \mathbb{R}^{N \times N}$ for each view $i$, where $(C_i)_{n,j}$ measures the similarity between the $n$-th rendering patch and the $j$-th RGB patch. 
For each rendering patch $f_{i, n}^r$, we aggregate its consistency with all RGB patches as
$c_{i,n} = \frac{1}{N} \sum_{j=1}^{N} (C_i)_{n,j}$.
Based on this score, we compute the local weight $w_{i,n} = 1 + \lambda \cdot \sigma(c_{i,n})$, where $\sigma(\cdot)$ is the sigmoid function and $\lambda$ controls the reweighting strength.
The reweighted local alignment loss is defined as:
\begin{equation}
\mathcal{L}_l^{w} = \frac{1}{V} \sum_{i=1}^{V} \frac{1}{N} \sum_{n=1}^{N} w_{i,n}
(1 - \langle f_{i,n}^{r_a},f_{i,n}^{R} \rangle).
\end{equation}

The overall reweighted alignment loss in this phase is defined as $\mathcal{L}_{\text{align}}^{w} = \mathcal{L}_g + \mathcal{L}_l^{w}$. 
Consequently, the reweighting strategy focuses more on regions with greater cross-modal discrepancies, as these regions contain richer semantic information that is critical for effective alignment.
By leveraging aggregated cross-modal semantic interactions, the proposed strategy adaptively emphasizes such regions during feature alignment, leading to more robust and discriminative representations.

\subsection{Modality-aware Prompt Learning}
\label{subsec:Modality-aware prompt learning}

Building upon the aligned features, we aim to learn anomaly semantics in a zero-shot manner. 
Accordingly, we propose a modality-aware prompt learning strategy that models anomaly representations for both RGB-aligned and rendering branches in separate semantic spaces.

\textbf{Dual-prompt learning.}
Following PointAD \cite{PointAD}, we construct learnable text prompts to model normal and anomalous semantics. 
Nevertheless, due to the discrepancy between RGB-aligned and rendering features, a shared prompt space is insufficient to capture modality-specific characteristics. 
We therefore introduce modality-specific prompts $t_m^N$ and $t_m^A$ for $m \in \{r_a, r\}$, corresponding to normal and anomalous states. 
These prompts are encoded into text embeddings $T_m^N$ and $T_m^A$ via a text encoder.
Based on text embeddings, the anomaly prediction for the $i$-th view is computed as
\[
\hat{y}_i^{m} = \frac{\exp(\langle G_i^{m}, T_m^A \rangle / \tau)}{\exp(\langle G_i^{m}, T_m^N \rangle / \tau) + \exp(\langle G_i^{m}, T_m^A \rangle / \tau)},
\]
where $\tau$ denotes the temperature coefficient. 
The final prediction $\hat{y}^{m}$ is obtained by averaging the predictions across all views.
For anomaly segmentation, we compute 2D anomaly score maps as
\[
M_{i, A}^{m} = 
\frac{\exp(\langle F_i^{m}, T_m^A \rangle)}
{\exp(\langle F_i^{m}, T_m^N \rangle) + \exp(\langle F_i^{m}, T_m^A \rangle)}.
\]

These maps are then back-projected to 3D space as
$M_{A}^{m} = \frac{1}{V} \sum_{i=1}^{V} R_i^{-1}(\mathrm{Up} (M_{i, A}^{m})) \odot H_i$, where $\mathrm{Up}(\cdot)$ denotes bilinear interpolation, $R_i^{-1}(\cdot)$ denotes the inverse projection from the $i$-th view to 3D space, $\odot$ denotes element-wise multiplication, and $H_i$ is the visibility mask for the corresponding view.
The same procedure is applied to obtain $M_{i, N}^{m}$ and $M_{N}^{m}$ for the normal state.

The supervision consists of image-level labels $y \in \{0,1\}$, point-level annotations $Y \in \{0,1\}^{P_n}$, and 2D masks $Y_i \in \{0,1\}^{H \times W}$. 
We jointly optimize classification and segmentation objectives for each modality $m$, denoted as $\mathcal{L}_{cls}^m$ and $\mathcal{L}_{seg}^m$, respectively. 
For image-level anomaly classification, we define the classification loss based on Cross-entropy (CE) as
\vspace{-5pt}
\[
\mathcal{L}_{cls}^m = \frac{1}{V} \sum_{i=1}^{V} \mathrm{CE}(\max\{Y_i\}, \hat{y}_i^m) + \mathrm{CE}(y, \hat{y}^m).
\] 

\vspace{-6pt}
For anomaly segmentation, we supervise both 3D point-level predictions and 2D pixel-level score maps, and the segmentation objectives at the 3D and 2D levels are defined using the Dice loss \cite{Dice_loss} and the Focal loss \cite{Focal_loss} as
\[
\mathcal{L}_{seg}^{m, 3D} = \mathrm{Dice}(M_A^m, Y) + \mathrm{Dice}(M_N^m, I - Y),
\]
\vspace{-6pt}
\vspace{-6pt}
\[
\mathcal{L}_{seg}^{m, 2D} = \frac{1}{V} \sum_{i=1}^V  \mathrm{Dice}(M_{i,A}^m, Y_i) + \mathrm{Dice}(M_{i,N}^m, I - Y_i) + \mathrm{Focal}(M_{i,N}^m \oplus M_{i,A}^m, Y_i),
\]
where $I$ is an all-one matrix and $\oplus$ denotes channel-wise concatenation.
Therefore, the overall segmentation loss is $\mathcal{L}_{seg}^m = \mathcal{L}_{seg}^{m, 3D} + \mathcal{L}_{seg}^{m, 2D}$, and the final objective is given by
$\mathcal{L} = \sum_{m \in \{r_a, r\}} \left( \mathcal{L}_{cls}^m + \mathcal{L}_{seg}^m \right)$.

\textbf{Dual-prompt contrastive alignment.}
Although modality-specific prompts are optimized independently for each branch, they still exhibit limited semantic expressiveness.
To address this limitation, we introduce a dual-prompt contrastive alignment that enhances prompt discriminability in the semantic space while preserving cross-modal consistency.

Concretely, for each modality $m \in \{r_a, r\}$, the prompts are optimized while treating the complementary modality $\bar{m}$ as a fixed reference. 
The contrastive alignment loss for modality $m$ is defined as:
\begin{equation}
\begin{aligned}
\mathcal{L}_{con}^{m}
= \frac{1}{2} \Bigg[
&- \log \frac{\exp(\langle T_m^N, T_{\bar{m}}^N\rangle)}
{\exp(\langle T_m^N, T_{\bar{m}}^N\rangle) + \exp(\langle T_m^N, T_m^A\rangle)} \\
&- \log \frac{\exp(\langle T_m^A, T_{\bar{m}}^A\rangle)}
{\exp( \langle T_m^A, T_{\bar{m}}^A\rangle) + \exp(\langle T_m^A, T_m^N\rangle)}
\Bigg],
\end{aligned}
\end{equation}
The overall objective in this phase is defined as
$
\mathcal{L} = \sum_{m \in \{r_a, r\}} \left( \mathcal{L}_{cls}^{m} + \mathcal{L}_{seg}^{m} + \lambda_{con} \mathcal{L}_{con}^{m} \right).
$
To ensure stable optimization, this regularization is activated in the later phase of modality-aware prompt learning, after the prompts have captured coarse normal and anomalous semantics.

\subsection{Training and Inference}

\textbf{Training.}
We adopt a two-stage training scheme with progressively refined objectives. 
In the first stage, the model is initially optimized using the alignment loss $\mathcal{L}_{align}$, 
and subsequently refined with the reweighted alignment loss $\mathcal{L}_{align}^{w}$  when the reweighting strategy is activated.
In the second stage, we jointly optimize classification and segmentation objectives $\mathcal{L} = \sum_{m \in \{r_a, r\}} \left( \mathcal{L}_{cls}^m + \mathcal{L}_{seg}^m \right)$ and further incorporate the contrastive loss $\mathcal{L}_{con}$ in the later phase of modality-aware prompt learning to refine text prompt  representations.

\textbf{Inference.}
Following Section \ref{subsec:Modality-aware prompt learning}, 3D anomaly score maps $M_A^m$ are computed for both modalities, yielding $M_A^{r_a}, M_A^{r}\in \mathbb{R}^{P_n}$ and corresponding image-level predictions $\hat{y}^{r_a}$ and $\hat{y}^{r}$.
The final 3D anomaly score map and image-level prediction are obtained via linear fusion $
M_A = \alpha M_A^{r_a} + (1 - \alpha) M_A^{r}, ~
\hat{y} = \alpha \hat{y}^{r_a} + (1 - \alpha)\hat{y}^{r} + \max(M_A),
$
where $\alpha \in [0,1]$ controls the contribution of each feature representation, and $\max(M_A)$ denotes the maximum anomaly score in $M_A$.

\section{Experiment}
\label{sec:experiment}

\subsection{Experiment Setup}
\textbf{Datasets and evaluation settings.} 
We evaluate our method under two settings: one-vs-rest and cross-dataset. 
In the one-vs-rest setting, experiments are conducted on MVTec3D-AD \cite{MVTec3D-AD} and Eyecandies \cite{Eyecandies}, where one object class is used as auxiliary training data and the remaining classes are used for evaluation. 
In the cross-dataset setting, the model is trained on a single category from MVTec3D-AD \cite{MVTec3D-AD} and evaluated on Eyecandies \cite{Eyecandies} and Real3D-AD \cite{Real3D-AD} to assess cross-dataset generalization. 
All experiments are repeated three times, and the average results are reported. 
More details on dataset descriptions, evaluation settings, and visualizations of the rendered multi-view observations are provided in Appendix \ref{sec:appendix_datasets}.

\textbf{Evaluation metrics.}
We adopt standard metrics to evaluate both classification and segmentation performance. 
For classification, we report the Area Under the Receiver Operating Characteristic curve (AUROC) and Average Precision (AP), denoted as O-R and O-A, respectively. 
For segmentation, we report AUROC and Per-Region Overlap (PRO), denoted as P-R and P-P, respectively.

\textbf{Implementation details.}
We adopt CLIP ViT-L/14@336px as the visual encoder. 
We generate 9 views for each object, where each view consists of both rendering and RGB observations. 
Our model is trained in two stages. 
In the first stage, we perform feature alignment for 250 epochs. 
The feature aligner consists of two separate MLPs, each with three layers of dimension 768 and GELU activation \cite{GELU}. 
The semantic consistency reweighting strategy is activated in the final 50 epochs.
In the second stage, we conduct modality-aware prompt learning for 15 epochs. 
The dual-prompt contrastive alignment is introduced in the final 5 epochs.
We use the Adam optimizer with a learning rate of $1 \times 10^{-3}$ for both stages. 
All experiments are conducted on NVIDIA RTX 4090 GPUs.
More details are provided in Appendix \ref{sec:appendix_implementation_details}.

\vspace{-6pt}
\begin{table}[h]
\centering
\begin{minipage}{0.52\linewidth}
\centering
\caption{One-vs-rest results on MVTec3D-AD and Eyecandies. Best results in \textbf{bold}.}
\resizebox{\linewidth}{!}{
\begin{tabular}{lcccccccc}
\hline
\multirow{2}{*}{Model} & \multicolumn{4}{c}{MVTec3D-AD} & \multicolumn{4}{c}{Eyecandies} \\ \cline{2-9}
& O-R & O-A & P-R & P-P & O-R & O-A & P-R & P-P \\ \hline
CLIP \cite{CLIP}            & 61.2 & 85.8 & 80.3 & 54.4 & 66.7 & 69.2 & 81.2 & 37.9 \\
AA-CLIP \cite{AA-CLIP}      & 74.2 & 88.5 & 87.1 & 61.6 & 66.5 & 68.0 & 85.9 & 39.3 \\
MVP-PCLP \cite{MVP-PCLIP}   & 81.3 & 92.7 & 94.6 & 83.6 & 69.3 & 72.7 & 90.8 & 67.8 \\
GS-CLIP* \cite{GS-CLIP}     & 78.7 & 92.9 & 94.6 & 80.6 & 68.3 & 73.4 & 92.3 & 72.0 \\
PointAD \cite{PointAD}      & 82.0 & 94.2 & 95.5 & 84.4 & 69.1 & 73.8 & 92.1 & 71.3 \\
ZUMA \cite{ZUMA}            & 81.9 & - & 95.6 & \textbf{87.0} & 67.1 & - & 92.5 & 73.6 \\
Align3D-AD & \textbf{83.0} & \textbf{94.3} & \textbf{95.9} & 86.5 & \textbf{71.3} & \textbf{76.6} & \textbf{92.6} & \textbf{74.2} \\ \hline
\end{tabular}
}
\label{tab:one_vs_rest}
\end{minipage}
\hfill
\begin{minipage}{0.44\linewidth}
\centering
\caption{Cross-dataset results on Eyecandies and Real3D-AD. Best results in \textbf{bold}.}
\resizebox{\linewidth}{!}{
\begin{tabular}{lccccccc}
\hline
 \multirow{2}{*}{Model} & \multicolumn{4}{c}{Eyecandies} & \multicolumn{3}{c}{Real3D-AD} \\ \cline{2-8}
& O-R & O-A & P-R & P-P & O-R & O-A & P-R \\ \hline
AA-CLIP \cite{AA-CLIP}      & 63.6 & 65.3 & 82.6 & 39.8 & 70.2 & 71.7 & 55.2 \\
MVP-PCLP \cite{MVP-PCLIP}   & 66.7 & 70.7 & 88.3 & 66.0 & 74.9 & 75.6 & 73.9 \\
GS-CLIP* \cite{GS-CLIP}     & 70.1 & 73.9 & 91.4 & 70.4 & 74.8 & 77.5 & 72.6 \\
PointAD \cite{PointAD}      & 69.5 & 74.3 & 91.8 & 71.4 & 75.9 & 77.9 & 71.6 \\
Align3D-AD & \textbf{70.7} & \textbf{75.3} & \textbf{92.0} & \textbf{72.6} & \textbf{76.1} & \textbf{79.3} & \textbf{76.7} \\ \hline
\end{tabular}
}
\label{tab:cross_dataset}
\end{minipage}
\end{table}

\begin{table}[ht]
\centering
\begin{minipage}{0.48\linewidth}
\centering
\caption{Ablation study of key components.
$^{*}$ denotes RGB-aligned feature representation only, and $^{\dagger}$ denotes rendering feature representation only.
}
\resizebox{\linewidth}{!}{
\begin{tabular}{cccccccc}
\hline
\multicolumn{4}{c}{Component} & \multicolumn{4}{c}{Metric} \\ \hline
DpL  & FA & SCR & DpCA & O-R & O-A & P-R & P-P \\ \hline
\ding{52} & \ding{55} & \ding{55}       & \ding{55} & 81.9 & 94.0 & 95.3 & 83.7 \\
\ding{52} & \ding{52}$^{*}$ & \ding{55} & \ding{55} & 79.8 & 93.0 & 95.6 & 84.2 \\
\ding{52} & \ding{52}$^{\dagger}$       & \ding{55} & \ding{55}  & 82.0 & 94.0 & 95.2 & 83.4 \\
\ding{52} & \ding{52} & \ding{55}       & \ding{55} & 82.4 & 94.0 & 95.7 & 85.6 \\
\ding{52} & \ding{52} & \ding{52}       & \ding{55} & 82.6 & 94.1 & 95.6 & 85.1 \\
\ding{52} & \ding{52} & \ding{55}       & \ding{52} & 82.6 & 94.2 & \textbf{95.9} & 86.3 \\
\ding{52} & \ding{52} & \ding{52}       & \ding{52} & \textbf{83.0} & \textbf{94.3} & \textbf{95.9} & \textbf{86.5} \\ \hline
\end{tabular}
}
\label{tab:Ablation studies of key components}
\end{minipage}
\hfill
\begin{minipage}{0.48\linewidth}
\centering
\caption{Ablation study of different feature representations.}
\resizebox{\linewidth}{!}{
\begin{tabular}{c c c c c c c c c}
\hline
\multirow{2}{*}{Modality}           & \multicolumn{4}{c}{MVTec3D-AD} & \multicolumn{4}{c}{Eyecandies} \\ \cline{2-9} 
 & O-R & O-A & P-R & P-P & O-R & O-A & P-R & P-P \\\hline
RGB-aligned     & 80.1 & 93.2 & 95.6 & 84.3 & 69.5 & 73.2 & 92.5 & 73.5 \\
Rendering       & 82.5 & 94.2 & 95.6 & 84.7 & 70.6 & 76.1 & 92.3 & 73.3 \\
Align3D-AD      & \textbf{83.0} & \textbf{94.3} & \textbf{95.9} & \textbf{86.5} & \textbf{71.3} & \textbf{76.6} & \textbf{92.6} & \textbf{74.2} \\
\hline
\end{tabular}
}
\label{tab:Ablation study of the modality}
\end{minipage}
\end{table}

\subsection{Main Results}

\textbf{One-vs-rest results.}
As shown in Table \ref{tab:one_vs_rest}, our method consistently outperforms 3D-based approaches. Compared with PointAD, which solely relies on rendering images, our method achieves improvements of +1.0\% in O-R, +0.1\% in O-A, +0.4\% in P-R, and +2.1\% in P-P on MVTec3D-AD, while also yielding notable gains on Eyecandies.
Compared with ZUMA, although our performance on MVTec3D-AD is slightly lower than ZUMA for the P-P metric, the overall performance remains superior, indicating stronger anomaly discrimination at both object and point levels.
The performance gains can be attributed to three key components. First, the introduction of RGB-aligned features effectively bridges the domain gap between rendering and RGB modalities. 
Second, the proposed dual-prompt learning strategy enables complementary modeling of anomaly semantics across modalities. 
Third, the contrastive alignment further facilitates effective modality-aware prompt learning by enforcing structured relationships between normal and anomalous representations. 
Together, these designs result in more accurate anomaly detection.

\begin{figure}[h]
\centering
\includegraphics[width=\linewidth]{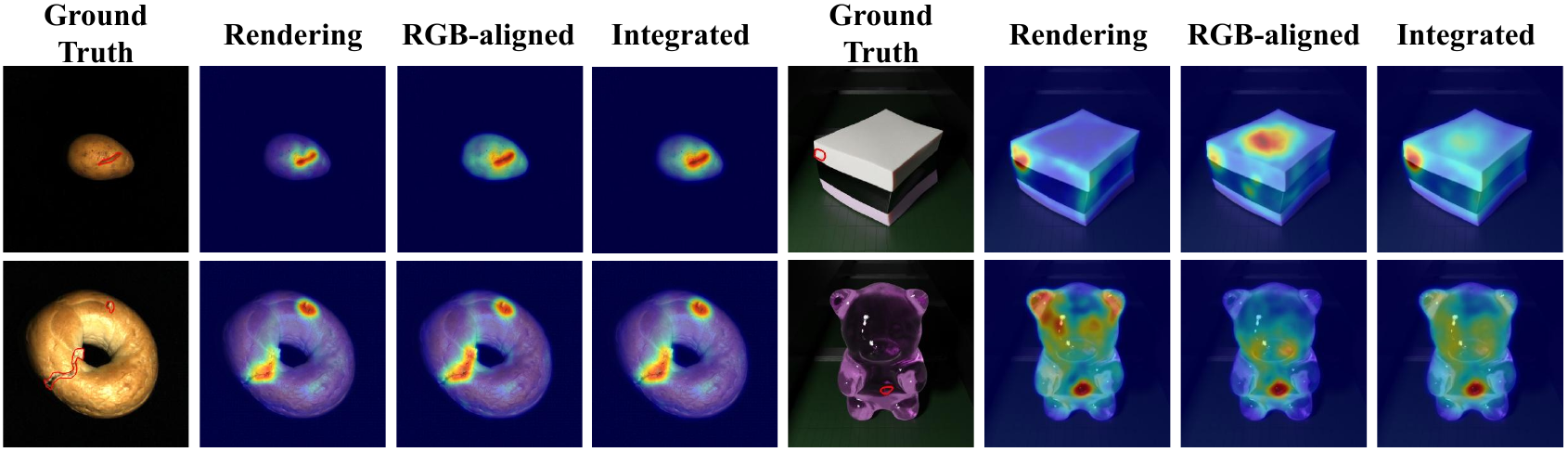}
\caption{Qualitative visualization on MVTec3D-AD and Eyecandies.}
\label{fig:Ablation study-Visualization}
\end{figure}

\textbf{Qualitative analysis.}
Figure \ref{fig:Ablation study-Visualization} presents qualitative comparisons among rendering, RGB-aligned, and integrated predictions, highlighting their complementary roles in anomaly detection. 
As shown in the first row, rendering features effectively capture geometric and structural information, enabling the detection of anomalies that are less distinguishable in the RGB-aligned space. 
In contrast, RGB-aligned features demonstrate a stronger advantage by leveraging richer semantic cues in other cases.
The second row shows that RGB-aligned features can also achieve more accurate localization, enabling more precise delineation of anomaly boundaries and reducing excessive erroneous segmentation, whereas rendering predictions are less discriminative.
This variation arises because rendering representations effectively encode geometric structures but have limited semantic expressiveness.
Therefore, although rendering features exhibit advantages for certain samples, incorporating RGB-aligned features yields superior overall performance and produces more accurate anomaly predictions than using rendering-only feature representations.

\textbf{Cross-dataset results.}
We further evaluate the zero-shot generalization capability of our model under the cross-dataset setting. 
As shown in Table \ref{tab:cross_dataset}, our method consistently outperforms existing approaches on both Eyecandies and Real3D-AD, demonstrating strong transferability across datasets with varying scene distributions. 
Methods relying solely on 3D information exhibit suboptimal generalization when encountering unseen semantics and appearance variations. 
In contrast, our method integrates semantic information aligned with RGB observations and geometric information gained from rendering representations, resulting in more robust anomaly detection across domains. 
Notably, even though Real3D-AD does not provide RGB data, our model still achieves strong performance, highlighting the importance of leveraging RGB-aligned representations to compensate for the limited semantic information.

\subsection{Ablation Study}
\label{subsec:ablation_study}
\textbf{Ablation of key components.}
To evaluate the effectiveness of each component, we conduct ablation studies on MVTec3D-AD by progressively enabling different modules, including dual-prompt learning (DpL), feature alignment (FA), semantic consistency reweighting (SCR), and dual-prompt contrastive alignment (DpCA), with results shown in Table \ref{tab:Ablation studies of key components}.
Under the setting with only DpL, the performance is relatively limited across all metrics. 
After introducing FA, overall performance is improved, with the integration of RGB-aligned and rendering features yielding more significant gains than using either feature representation alone.
Building upon the integrated setting, SCR further enhances feature representations by emphasizing holistic semantic consistency, while DpCA improves the discriminability of prompt representations.
With all components enabled, the model achieves the best performance, validating the effectiveness of our design.

\textbf{Feature representation analysis.}
To evaluate the effectiveness of feature alignment, we compare three configurations: using RGB-aligned representations, rendering representations, and integrated representations on MVTec3D-AD and Eyecandies. 
As shown in Table \ref{tab:Ablation study of the modality}, relying on a single representation source leads to suboptimal performance. 
Compared to rendering features, RGB-aligned features achieve slightly lower performance while providing complementary semantic information. 
Nevertheless, incorporating RGB-aligned features consistently improves performance over using rendering features alone across all metrics on both datasets.
These results demonstrate that RGB-aligned and rendering features are complementary, highlighting the effectiveness of the proposed cross-modal feature alignment strategy in bridging the domain gap.
We provide more analysis in Appendix \ref{sec:appendix_analysis_of_RGB-aligned_and_rendering_representations} and Appendix \ref{sec:appendix_analysis_of_semantic_consistency_reweighting} to further evaluate the effectiveness of RGB-aligned representations and illustrate how the semantic consistency reweighting strategy improves the cross-modal feature alignment for learning more effective RGB-aligned feature representations.

\textbf{Impact of optimization epochs for SCR and DpCA.}
Table \ref{tab:ablation_study_epoch_of_stage 1.2} and Table \ref{tab:ablation_study_epoch_of_stage 2.2} present the impact of different optimization epochs for SCR and DpCA, respectively. 
As shown in these tables, the performance varies with the number of optimization epochs for both SCR and DpCA. 
For SCR, performance is suboptimal with a small number of optimization epochs, reaches its peak at 50 epochs, and then slightly degrades with additional optimization epochs, as it tends to overemphasize semantic information and suppress geometric cues, leading to excessive alignment. 
Similarly, for DpCA, performance peaks at 5 epochs and gradually declines with further optimization, which can be attributed to the prompt space becoming overly constrained under prolonged training.
These results suggest that an appropriate number of optimization epochs is critical, as insufficient training limits effectiveness while excessive optimization may harm generalization. 
Nevertheless, the performance remains relatively stable under different optimization epochs, indicating the robustness of our model.

\begin{table}[h]
\centering
\begin{minipage}{0.45\linewidth}
\centering
\small
\caption{Impact of SCR optimization epochs.}
\begin{tabular}{ccccc}
\hline
Optimization Epochs & O-R & O-A & P-R & P-P \\ \hline
10  & 82.4 & 94.1 & 95.7 & 85.6 \\
30  & 82.6 & 94.1 & 95.5 & 85.0 \\
50  & \textbf{83.0} & \textbf{94.3} & \textbf{95.9} & \textbf{86.5} \\
100 & 82.5 & 94.1 & 95.6 & 85.3 \\
200 & 82.4 & 94.1 & 95.4 & 84.6 \\
250 & 82.4 & 94.1 & 95.8 & 86.0 \\ \hline
\end{tabular}
\label{tab:ablation_study_epoch_of_stage 1.2}
\end{minipage}
\hfill
\begin{minipage}{0.48\linewidth}
\centering
\small
\caption{Impact of DpCA optimization epochs.}
\begin{tabular}{ccccc}
\hline
Optimization Epochs & O-R & O-A & P-R & P-P \\ \hline
1  & 82.6 & 94.2 & 95.7 & 85.7 \\
3  & 82.6 & 94.1 & 95.7 & 85.7 \\
5  & \textbf{83.0} & \textbf{94.3} & \textbf{95.9} & \textbf{86.5} \\
8  & 82.7 & 94.2 & 95.7 & 85.9 \\
10 & 82.5 & 94.1 & 95.7 & 85.8 \\
15 & 82.2 & 94.0 & 95.3 & 84.1 \\ \hline
\end{tabular}
\label{tab:ablation_study_epoch_of_stage 2.2}
\end{minipage}
\end{table}

\begin{wraptable}{r}{0.45\linewidth}
\vspace{-10pt}
\centering
\caption{Comparison of inference time per image, frame per second (FPS), and GPU memory usage on MVTec3D-AD.}
\resizebox{0.95\linewidth}{!}{
\begin{tabular}{lccc}
\toprule
Model & Time (s) & FPS & Memory (MB) \\
\midrule
GS-CLIP* & 0.62 & 1.62 & 4978 \\
PointAD* & 0.18 & 5.68 & 2460 \\
Ours & 0.19 & 5.30 & 2791 \\
\bottomrule
\end{tabular}
}
\label{tab:ablation_study_complexity_analysis}
\vspace{-10pt}
\end{wraptable}

\textbf{Complexity analysis.}
We investigate the computational efficiency of mainstream methods, including PointAD and GS-CLIP, in terms of inference time per image, frames per second (FPS), and GPU memory usage, with results reported in Table \ref{tab:ablation_study_complexity_analysis}. 
Compared with PointAD, which relies on rendering images, our method exhibits slightly higher inference time per image and GPU memory usage, while exhibiting a slightly lower FPS.
Compared with GS-CLIP, which utilizes point clouds, rendering images, and depth maps, our method achieves consistently better efficiency. 
Overall, our method obtains the best performance, achieving a more favorable trade-off between computational cost and accuracy.
More details are provided in Appendix \ref{sec:appendix_implementation_details_for_complexity_analysis}.

\section{Conclusion}
\label{sec:conclusion}
This paper studies the challenging problem of zero-shot 3D anomaly detection. 
We propose Align3D-AD, a unified framework that bridges rendering and RGB modalities to learn transferable anomaly representations.  
By introducing cross-modal feature alignment, the model transfers semantic knowledge from RGB observations to rendering features. 
A semantic consistency reweighting strategy is further employed to enhance cross-modal feature alignment by leveraging holistic semantic consistency.
Furthermore, a modality-aware prompt learning scheme with dual-prompt contrastive alignment is developed to capture modality-specific anomaly semantics and enhance the discriminability of prompt representations. 
Extensive experiments on multiple benchmarks demonstrate that our method achieves superior performance under both one-vs-rest and cross-dataset settings, highlighting its strong generalization capability and effectiveness.

\textbf{Limitations.}
In practice, introducing RGB guidance increases model complexity due to the additional modality and modality-specific prompt learners. 
Moreover, the reliance on multi-view observations makes the model sensitive to rendering quality, which may affect performance.

\textbf{Broader impact.}
This work takes an initial step toward bridging the domain gap in zero-shot 3D anomaly detection by explicitly incorporating RGB information into the training stage.
We hope this perspective can encourage future research to move beyond purely 3D information and focus more on cross-modal understanding for 3D tasks. 


\bibliographystyle{plainnat} 
\bibliography{reference}

@article{PointAD,
  title={Pointad: Comprehending 3d anomalies from points and pixels for zero-shot 3d anomaly detection},
  author={Zhou, Qihang and Yan, Jiangtao and He, Shibo and Meng, Wenchao and Chen, Jiming},
  journal={Advances in Neural Information Processing Systems},
  volume={37},
  pages={84866--84896},
  year={2024}
}

@article{MVTec3D-AD,
  title={The mvtec 3d-ad dataset for unsupervised 3d anomaly detection and localization},
  author={Bergmann, Paul and Jin, Xin and Sattlegger, David and Steger, Carsten},
  journal={arXiv preprint arXiv:2112.09045},
  year={2021}
}

@inproceedings{Eyecandies,
  title={The eyecandies dataset for unsupervised multimodal anomaly detection and localization},
  author={Bonfiglioli, Luca and Toschi, Marco and Silvestri, Davide and Fioraio, Nicola and De Gregorio, Daniele},
  booktitle={Proceedings of the Asian Conference on Computer Vision},
  pages={3586--3602},
  year={2022}
}

@article{Real3D-AD,
  title={Real3d-ad: A dataset of point cloud anomaly detection},
  author={Liu, Jiaqi and Xie, Guoyang and Chen, Ruitao and Li, Xinpeng and Wang, Jinbao and Liu, Yong and Wang, Chengjie and Zheng, Feng},
  journal={Advances in Neural Information Processing Systems},
  volume={36},
  pages={30402--30415},
  year={2023}
}

@article{GS-CLIP,
  title={GS-CLIP: Zero-shot 3D anomaly detection by geometry-aware prompt and synergistic view representation learning},
  author={Deng, Zehao and Liu, An and Wang, Yan},
  journal={arXiv preprint arXiv:2602.19206},
  year={2026}
}

@ARTICLE{ZUMA,
  author={Ma, Yunfeng and Liu, Min and Jiang, Shuai and Zhou, Jingyu and Bian, Yuan and Wang, Xueping and Wang, Yaonan},
  journal={IEEE Transactions on Pattern Analysis and Machine Intelligence}, 
  title={ZUMA: Training-free zero-shot unified multimodal anomaly detection}, 
  year={2026},
  volume={},
  number={},
  pages={1-14},
  doi={10.1109/TPAMI.2026.3658856}
}

@ARTICLE{MVP-PCLIP,
  author={Cheng, Yuqi and Cao, Yunkang and Xie, Guoyang and Lu, Zhichao and Shen, Weiming},
  journal={IEEE Transactions on Systems, Man, and Cybernetics: Systems}, 
  title={Toward zero-shot point cloud anomaly detection: a multiview projection framework}, 
  year={2026},
  volume={56},
  number={3},
  pages={1747-1760},
  doi={10.1109/TSMC.2025.3648581}
}

@inproceedings{AA-CLIP,
  title={Aa-clip: Enhancing zero-shot anomaly detection via anomaly-aware clip},
  author={Ma, Wenxin and Zhang, Xu and Yao, Qingsong and Tang, Fenghe and Wu, Chenxu and Li, Yingtai and Yan, Rui and Jiang, Zihang and Zhou, S Kevin},
  booktitle={Proceedings of the Computer Vision and Pattern Recognition Conference},
  pages={4744--4754},
  year={2025}
}

@inproceedings{CLIP,
  title={Learning transferable visual models from natural language supervision},
  author={Radford, Alec and Kim, Jong Wook and Hallacy, Chris and Ramesh, Aditya and Goh, Gabriel and Agarwal, Sandhini and Sastry, Girish and Askell, Amanda and Mishkin, Pamela and Clark, Jack and others},
  booktitle={International Conference on Machine Learning},
  pages={8748--8763},
  year={2021},
  organization={PmLR}
}

@article{du20253d,
  title={3D vision-based anomaly detection in manufacturing: A survey},
  author={Du, Juan and Tao, Chengyu and Cao, Xuanming and Tsung, Fugee},
  journal={Frontiers of Engineering Management},
  volume={12},
  number={2},
  pages={343--360},
  year={2025},
  publisher={Springer}
}

@article{tao2025pointsgrade,
  title={PointSGRADE: Sparse learning with graph representation for anomaly detection by using unstructured 3D point cloud data},
  author={Tao, Chengyu and Du, Juan},
  journal={IISE Transactions},
  volume={57},
  number={2},
  pages={131--144},
  year={2025},
  publisher={Taylor \& Francis}
}

@article{cao2025iaenet,
  title={IAENet: An importance-aware ensemble model for 3D point cloud-based anomaly detection},
  author={Cao, Xuanming and Tao, Chengyu and Cheng, Yifeng and Du, Juan},
  journal={Information Fusion},
  pages={104097},
  year={2025},
  publisher={Elsevier}
}

@inproceedings{horwitz2023back,
  title={Back to the feature: classical 3d features are (almost) all you need for 3d anomaly detection},
  author={Horwitz, Eliahu and Hoshen, Yedid},
  booktitle={Proceedings of the IEEE/CVF Conference on Computer Vision and Pattern Recognition},
  pages={2968--2977},
  year={2023}
}

@inproceedings{roth2022towards,
  title={Towards total recall in industrial anomaly detection},
  author={Roth, Karsten and Pemula, Latha and Zepeda, Joaquin and Sch{\"o}lkopf, Bernhard and Brox, Thomas and Gehler, Peter},
  booktitle={Proceedings of the IEEE/CVF Conference on Computer Vision and Pattern Recognition},
  pages={14318--14328},
  year={2022}
}

@inproceedings{bergmann2023anomaly,
  title={Anomaly detection in 3d point clouds using deep geometric descriptors},
  author={Bergmann, Paul and Sattlegger, David},
  booktitle={Proceedings of the IEEE/CVF Winter Conference on Applications of Computer Vision},
  pages={2613--2623},
  year={2023}
}

@inproceedings{zhou2023anomalyclip,
  title={AnomalyCLIP: Object-agnostic prompt learning for zero-shot anomaly detection},
  author={Zhou, Qihang and Pang, Guansong and Tian, Yu and He, Shibo and Chen, Jiming},
  booktitle={The Twelfth International Conference on Learning Representations},
  year={2023}
}

@inproceedings{wang2023multimodal,
  title={Multimodal industrial anomaly detection via hybrid fusion},
  author={Wang, Yue and Peng, Jinlong and Zhang, Jiangning and Yi, Ran and Wang, Yabiao and Wang, Chengjie},
  booktitle={Proceedings of the IEEE/CVF Conference on Computer Vision and Pattern Recognition},
  pages={8032--8041},
  year={2023}
}

@inproceedings{jeong2023winclip,
  title={Winclip: Zero-/few-shot anomaly classification and segmentation},
  author={Jeong, Jongheon and Zou, Yang and Kim, Taewan and Zhang, Dongqing and Ravichandran, Avinash and Dabeer, Onkar},
  booktitle={Proceedings of the IEEE/CVF Conference on Computer Vision and Pattern Recognition},
  pages={19606--19616},
  year={2023}
}

@inproceedings{cao2024adaclip,
  title={Adaclip: Adapting clip with hybrid learnable prompts for zero-shot anomaly detection},
  author={Cao, Yunkang and Zhang, Jiangning and Frittoli, Luca and Cheng, Yuqi and Shen, Weiming and Boracchi, Giacomo},
  booktitle={European Conference on Computer Vision},
  pages={55--72},
  year={2024},
  organization={Springer}
}

@inproceedings{qu2025bayesian,
  title={Bayesian prompt flow learning for zero-shot anomaly detection},
  author={Qu, Zhen and Tao, Xian and Gong, Xinyi and Qu, Shichen and Chen, Qiyu and Zhang, Zhengtao and Wang, Xingang and Ding, Guiguang},
  booktitle={Proceedings of the IEEE/CVF Conference on Computer Vision and Pattern Recognition},
  pages={30398--30408},
  year={2025}
}

@article{cao2024complementary,
  title={Complementary pseudo multimodal feature for point cloud anomaly detection},
  author={Cao, Yunkang and Xu, Xiaohao and Shen, Weiming},
  journal={Pattern Recognition},
  volume={156},
  pages={110761},
  year={2024},
  publisher={Elsevier}
}

@inproceedings{rudolph2023asymmetric,
  title={Asymmetric student-teacher networks for industrial anomaly detection},
  author={Rudolph, Marco and Wehrbein, Tom and Rosenhahn, Bodo and Wandt, Bastian},
  booktitle={Proceedings of the IEEE/CVF Winter Conference on Applications of Computer Vision},
  pages={2592--2602},
  year={2023}
}

@inproceedings{li2024towards,
  title={Towards scalable 3d anomaly detection and localization: A benchmark via 3d anomaly synthesis and a self-supervised learning network},
  author={Li, Wenqiao and Xu, Xiaohao and Gu, Yao and Zheng, Bozhong and Gao, Shenghua and Wu, Yingna},
  booktitle={Proceedings of the IEEE/CVF Conference on Computer Vision and Pattern Recognition},
  pages={22207--22216},
  year={2024}
}

@inproceedings{zavrtanik2024cheating,
  title={Cheating depth: Enhancing 3d surface anomaly detection via depth simulation},
  author={Zavrtanik, Vitjan and Kristan, Matej and Sko{\v{c}}aj, Danijel},
  booktitle={Proceedings of the IEEE/CVF Winter Conference on Applications of Computer Vision},
  pages={2164--2172},
  year={2024}
}

@InProceedings{chu2023shape,
  title = {Shape-Guided dual-memory learning for 3D anomaly detection},
  author = {Chu, Yu-Min and Liu, Chieh and Hsieh, Ting-I and Chen, Hwann-Tzong and Liu, Tyng-Luh},
  booktitle = {Proceedings of the 40th International Conference on Machine Learning},
  pages = {6185--6194},
  year = {2023},
}

@inproceedings{gu2024filo,
  title={Filo: Zero-shot anomaly detection by fine-grained description and high-quality localization},
  author={Gu, Zhaopeng and Zhu, Bingke and Zhu, Guibo and Chen, Yingying and Li, Hao and Tang, Ming and Wang, Jinqiao},
  booktitle={Proceedings of the 32nd ACM International Conference on Multimedia},
  pages={2041--2049},
  year={2024}
}

@inproceedings{gu2024anomalygpt,
  title={Anomalygpt: Detecting industrial anomalies using large vision-language models},
  author={Gu, Zhaopeng and Zhu, Bingke and Zhu, Guibo and Chen, Yingying and Tang, Ming and Wang, Jinqiao},
  booktitle={Proceedings of the AAAI Conference on Artificial Intelligence},
  volume={38},
  number={3},
  pages={1932--1940},
  year={2024}
}

@inproceedings{G2SF,
  title={G2SF: Geometry-guided score fusion for multimodal industrial anomaly detection},
  author={Tao, Chengyu and Cao, Xuanming and Du, Juan},
  booktitle={Proceedings of the IEEE/CVF International Conference on Computer Vision},
  pages={20551--20560},
  year={2025}
}

@inproceedings{cao2023anomaly,
  title={Anomaly detection under distribution shift},
  author={Cao, Tri and Zhu, Jiawen and Pang, Guansong},
  booktitle={Proceedings of the IEEE/CVF International Conference on Computer Vision},
  pages={6511--6523},
  year={2023}
}

@inproceedings{bergmann2020uninformed,
  title={Uninformed students: Student-teacher anomaly detection with discriminative latent embeddings},
  author={Bergmann, Paul and Fauser, Michael and Sattlegger, David and Steger, Carsten},
  booktitle={Proceedings of the IEEE/CVF Conference on Computer Vision and Pattern Recognition},
  pages={4183--4192},
  year={2020}
}

@article{cohen2020sub,
  title={Sub-image anomaly detection with deep pyramid correspondences},
  author={Cohen, Niv and Hoshen, Yedid},
  journal={arXiv preprint arXiv:2005.02357},
  year={2020}
}

@article{GELU,
  title={Gaussian error linear units (gelus)},
  author={Hendrycks, Dan and Gimpel, Kevin},
  journal={arXiv preprint arXiv:1606.08415},
  year={2016}
}

@inproceedings{Focal_loss,
  title={Focal loss for dense object detection},
  author={Lin, Tsung-Yi and Goyal, Priya and Girshick, Ross and He, Kaiming and Doll{\'a}r, Piotr},
  booktitle={Proceedings of the IEEE International Conference on Computer Vision},
  pages={2980--2988},
  year={2017}
}

@inproceedings{Dice_loss,
  title={V-net: Fully convolutional neural networks for volumetric medical image segmentation},
  author={Milletari, Fausto and Navab, Nassir and Ahmadi, Seyed-Ahmad},
  booktitle={2016 Fourth International Conference on 3D Vision (3DV)},
  pages={565--571},
  year={2016},
  organization={Ieee}
}


\appendix

\section{Dataset}
\label{sec:appendix_datasets}

\textbf{Dataset choice and description.}
We conduct experiments on three publicly available 3D anomaly detection datasets: MVTec3D-AD, Eyecandies, and Real3D-AD. 
MVTec3D-AD contains high-resolution 3D scans from 10 object categories with diverse semantics and provides both point-level annotations and corresponding RGB images.
Eyecandies is a synthetic dataset with 10 categories, also offering paired RGB observations. 
Real3D-AD is a real-world industrial dataset covering 12 categories, containing both normal and defective instances but without RGB information.

Since our framework leverages RGB information during training, we use MVTec3D-AD and Eyecandies for training and evaluation under the one-vs-rest setting. 
In addition, we consider a more challenging cross-dataset setting, where auxiliary data from one dataset is used to evaluate all categories in another dataset. 
Specifically, the model is trained on MVTec3D-AD and evaluated on Eyecandies and Real3D-AD. 
Notably, despite the absence of RGB data in Real3D-AD, our model still generalizes effectively, as the cross-modal alignment learned during training enables rendering features to produce RGB-aligned representations, thereby compensating for the limited semantic expressiveness of 3D data.

\textbf{Evaluation settings.}
Under the one-vs-rest setting, we report the average performance by alternately selecting different object categories as the auxiliary training set, including \emph{Carrot}, \emph{Cookie}, and \emph{Dowel} for MVTec3D-AD, and \emph{Confetto}, \emph{Licorice Sandwich}, and \emph{Peppermint Candy} for Eyecandies, with the average performance reported. 

For the cross-dataset setting, the model is trained on the \emph{Carrot}, \emph{Cookie}, and \emph{Dowel} categories from MVTec3D-AD and directly evaluated on all categories of Eyecandies and Real3D-AD, with the average performance reported.

\begin{figure}[t]
\centering
\includegraphics[width=\linewidth]{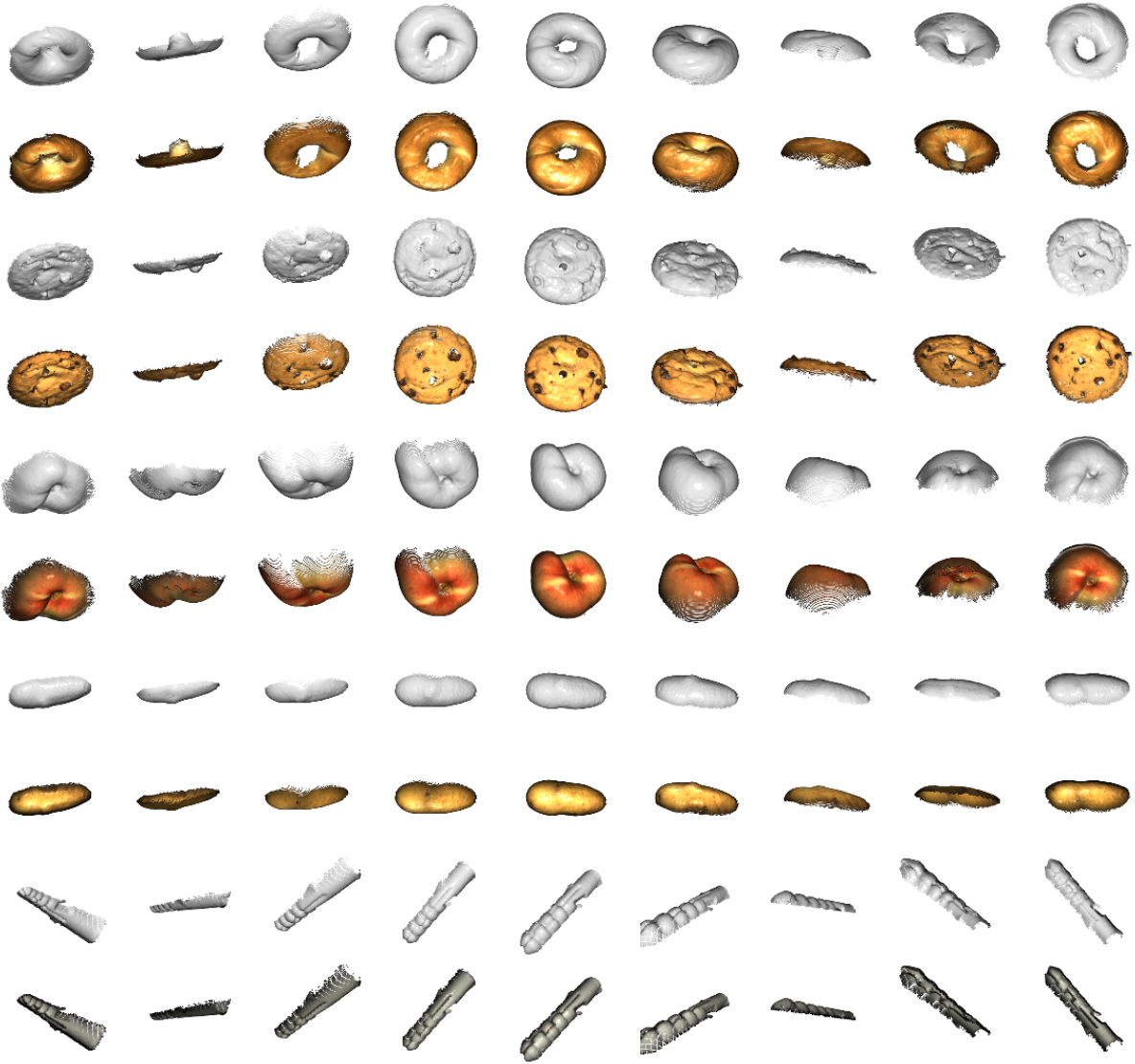}
\caption{Visualization of the rendered multi-view observations on MVTec3D-AD, including both rendering and corresponding RGB images from different views.}
\label{fig:Appendix_MVTec_Dataset_Visualization}
\end{figure}

\begin{figure}[t]
\centering
\includegraphics[width=\linewidth]{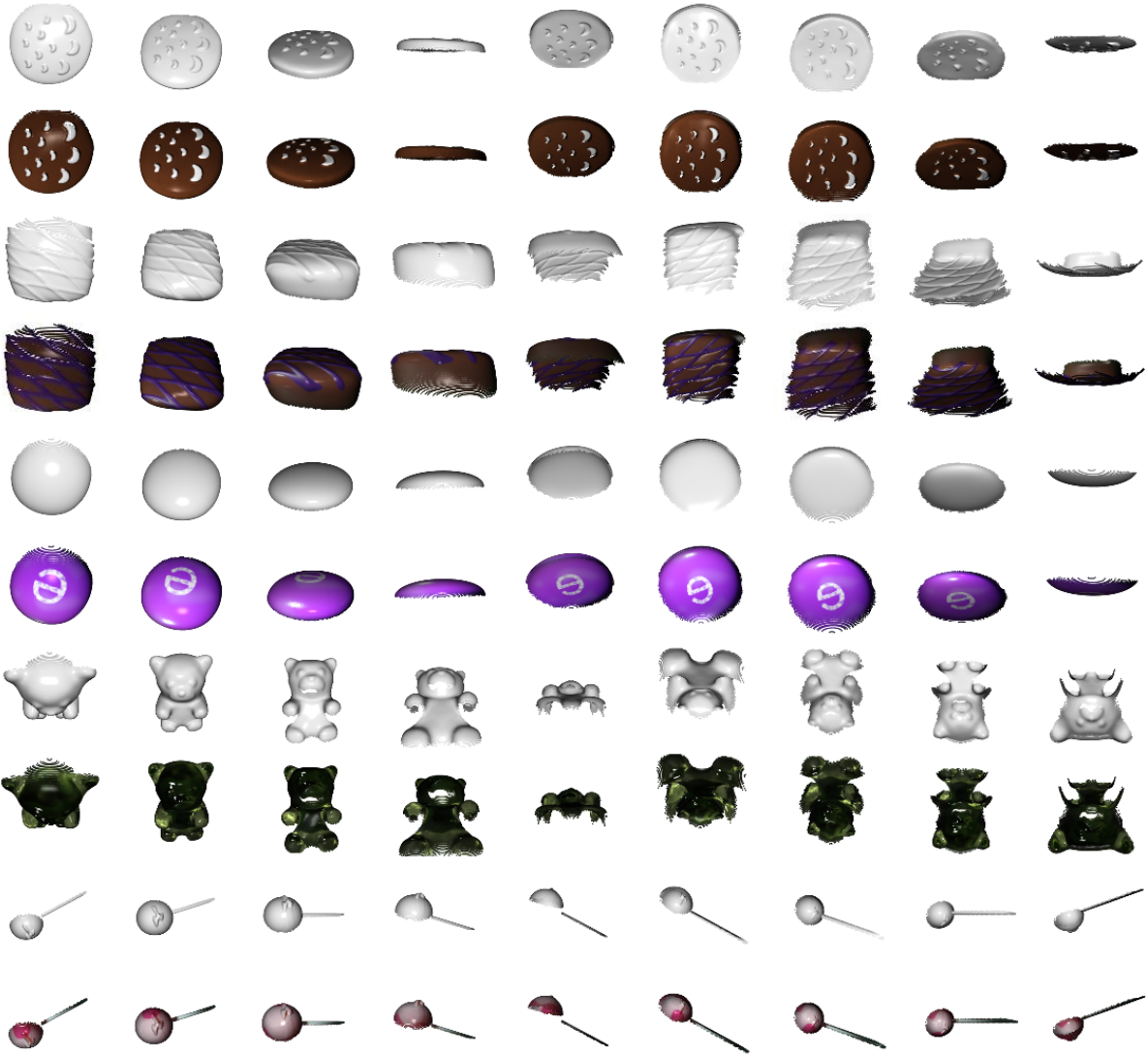}
\caption{Visualization of the rendered multi-view observations on Eyecandies, including both rendering and corresponding RGB images from different views.}
\label{fig:Appendix_Eyecandies_Dataset_Visualization}
\end{figure}

\textbf{Visualization of the rendered multi-view datasets.}
We present visualizations of the rendered observations used in our framework to illustrate the characteristics of the multi-view observations, as shown in Figure \ref{fig:Appendix_MVTec_Dataset_Visualization} for MVTec3D-AD and Figure \ref{fig:Appendix_Eyecandies_Dataset_Visualization} for Eyecandies.
For each 3D point cloud sample, rendering images along with corresponding RGB observations are generated from diverse viewpoints. 
From these visualizations, it can be observed that rendering images primarily capture geometric structures and shape information, whereas RGB observations provide richer semantic and appearance cues. 
By leveraging this complementary information, Align3D-AD constructs RGB-aligned representations to bridge the domain gap between rendering and RGB modalities, leading to more discriminative and robust anomaly detection.

However, due to the inherent characteristics of the collected point cloud data, the multi-view observations obtained through the rendering process may exhibit missing regions, blank areas, and noise artifacts. 
These imperfections introduce additional challenges for both cross-modal feature alignment and zero-shot 3D anomaly detection, as establishing reliable correspondences across modalities becomes more difficult under incomplete and noisy observations. 
Further analysis is provided in Appendix \ref{sec:appendix_failure_cases}.

\section{Implementation Details}
\label{sec:appendix_implementation_details}
\vspace{-6pt}
We adopt CLIP ViT-L/14@336px as the visual encoder with pre-trained weights and keep it frozen throughout training. 
For each 3D object, we use the Open3D library to generate 9 views by rotating point clouds around the X-axis with angles $\left\{-\frac{4\pi}{5}, -\frac{3\pi}{5}, -\frac{2\pi}{5}, -\frac{\pi}{5}, 0, \frac{\pi}{5}, \frac{2\pi}{5}, \frac{3\pi}{5}, \frac{4\pi}{5}\right\}$ for all categories. 
Correspondingly, the 3D point-level annotations are projected to obtain 2D masks for each view, following the standard rendering and projection process as PointAD \cite{PointAD}.
All images are resized to $336 \times 336$.

In the first stage, we train the feature aligner for 250 epochs to project rendering features into the RGB feature space. 
The feature aligner consists of two independent MLPs for global and local features, each comprising three fully connected layers with a hidden dimension of 768 and GELU activation. 
To improve efficiency, image features are pre-computed and cached before training. 
In the final 50 epochs, the semantic consistency reweighting strategy is introduced to refine feature local alignment.
In the second stage, we optimize modality-specific prompt learners for 15 epochs while keeping the feature aligner frozen.  
In the final 5 epochs, the dual-prompt contrastive alignment is incorporated to enhance prompt discriminability while preserving cross-modal consistency.

The modality fusion coefficient $\alpha$ during inference is set to 0.5, the reweighting parameter $\lambda$ in SCR is set to 1, the contrastive alignment loss weight $\lambda_{con}$ in DpCA is set to 0.05, and the temperature coefficient $\tau$ is set to 0.07 for most categories.
It is worth noting that these parameters may require fine-tuning for specific categories to balance distributions across modalities and ensure stable convergence in zero-shot 3D anomaly detection. 
All learnable components are optimized using the Adam optimizer with a learning rate of $1 \times 10^{-3}$. 
During the training stage, the batch size is set to 4.  
All experiments are implemented using PyTorch 2.7.1 and executed on NVIDIA RTX 4090 GPUs.

Note that since RGB information is only introduced in stage 1 to facilitate cross-modal feature alignment and reduce the domain gap between RGB and rendering modalities, while stage 2 and inference rely solely on rendering images, our method remains consistent with the zero-shot 3D anomaly detection setting.

\section{Analysis of RGB-aligned Feature Representations}
\label{sec:appendix_analysis_of_RGB-aligned_and_rendering_representations}
In this section, we investigate the role of RGB-aligned feature representations, with results under different auxiliary training categories reported in Table \ref{tab:appendix_modality_analysis_mvtec} for MVTec3D-AD and Table \ref{tab:appendix_modality_analysis_eyecandies} for Eyecandies. 
The leftmost column corresponds to different auxiliary training categories used under the one-vs-rest setting.
According to Table \ref{tab:appendix_modality_analysis_mvtec} and Table \ref{tab:appendix_modality_analysis_eyecandies}, 
although RGB-aligned features are slightly inferior to rendering features,
they still achieve competitive overall performance, indicating the ability to capture rich semantic information complementary to geometric cues.

We further analyze the segmentation performance. 
As reported in Table \ref{tab:Ablation study of the modality} in Section \ref{subsec:ablation_study}, RGB-aligned features achieve comparable performance to rendering features on MVTec3D-AD, while outperforming rendering features on Eyecandies.
Moreover, as shown in Table \ref{tab:appendix_modality_analysis_mvtec} and Table \ref{tab:appendix_modality_analysis_eyecandies}, for both the P-R and P-P metrics under different auxiliary training categories, RGB-aligned features outperform rendering features in most auxiliary training settings, including \emph{Carrot} and \emph{Cookie} on MVTec3D-AD, as well as \emph{Confetto} and \emph{Licorice Sandwich} on Eyecandies.
These results suggest that RGB-aligned representations provide complementary semantic cues that are beneficial for anomaly localization.
Although the overall performance of RGB-aligned features is slightly inferior to rendering features, this gap is primarily driven by specific auxiliary training categories, namely \emph{Dowel} on MVTec3D-AD and \emph{Peppermint Candy} on Eyecandies.

These observations further demonstrate that, even without RGB inputs during inference, the learned feature aligner enables the model to extract semantically enriched representations from rendering observations, validating the effectiveness of the proposed cross-modal feature alignment.

\begin{table}[]
\centering
\caption{Comparison of RGB-aligned and rendering feature representations under different auxiliary training categories on MVTec3D-AD under the one-vs-rest setting.}
\begin{tabular}{ccccccccc}
\hline
\multirow{2}{*}{Category} & \multicolumn{4}{c}{RGB-aligned}                  & \multicolumn{4}{c}{Rendering}                    \\ \cline{2-9} 
       & O-R & O-A & P-R & P-P & O-R & O-A & P-R & P-P \\ \hline
Carrot & 79.8         & 93.0         & 96.0         & 85.8         & 79.6         & 93.4         & 95.6         & 84.5         \\
Cookie & 77.0         & 92.4         & 96.9         & 87.0         & 82.2         & 94.1         & 96.4         & 85.3         \\
Dowel  & 81.6         & 93.2         & 94.3         & 80.7         & 84.2         & 94.4         & 95.3         & 84.8         \\ \hline
\end{tabular}
\label{tab:appendix_modality_analysis_mvtec}
\end{table}

\begin{table}[]
\centering
\caption{Comparison of RGB-aligned and rendering feature representations under different auxiliary training categories on Eyecandies under the one-vs-rest setting.}
\begin{tabular}{ccccccccc}
\hline
\multirow{2}{*}{Category} & \multicolumn{4}{c}{RGB-aligned}                  & \multicolumn{4}{c}{Rendering}                    \\ \cline{2-9}
& O-R & O-A & P-R & P-P & O-R & O-A & P-R & P-P \\ \hline
Confetto          & 67.6         & 70.5         & 93.0         & 74.6         & 69.3         & 74.2         & 92.3         & 73.7         \\
Licorice Sandwich & 66.5         & 71.5         & 92.0         & 73.1         & 69.3         & 75.4         & 91.6         & 71.1         \\
Peppermint Candy  & 71.1         & 73.6         & 92.1         & 71.8         & 70.9         & 75.9         & 92.6         & 74.3         \\ \hline
\end{tabular}
\label{tab:appendix_modality_analysis_eyecandies}
\end{table}

\section{Analysis of Semantic Consistency Reweighting}
\label{sec:appendix_analysis_of_semantic_consistency_reweighting}
\begin{figure}[h]
\centering
\includegraphics[width=\linewidth]{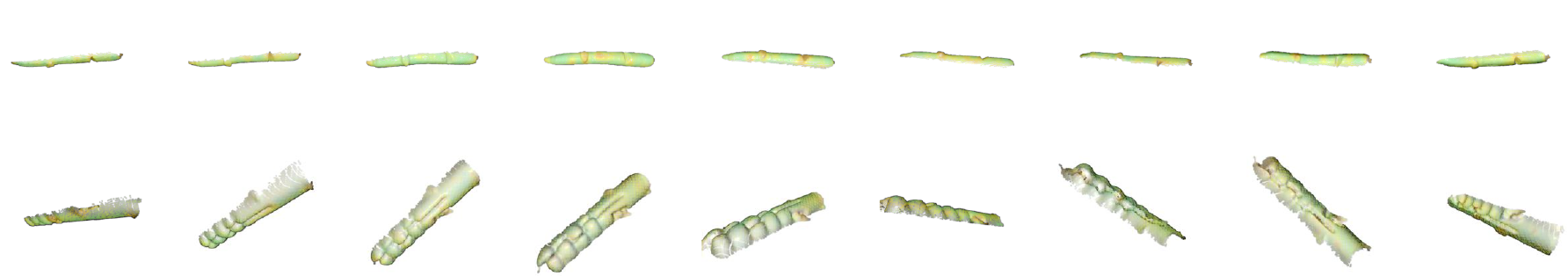}
\caption{Visualization of semantic consistency weights on rendering images. Higher weights are represented in red, whereas lower weights are represented in green.}
\label{fig:Appendix_SCR_visualization}
\end{figure}

To better understand the behavior of the semantic consistency reweighting strategy, we visualize the local weight by overlaying it onto the rendering images, where higher weights are represented in red and lower weights are represented in green.
As shown in Figure \ref{fig:Appendix_SCR_visualization}, high weights are primarily concentrated in geometrically deformed and structurally complex regions. 
This spatial pattern indicates that the weighting mechanism assigns different importance to regions with varying geometric and semantic characteristics.

This phenomenon arises because regions with structural deformation and complexity typically contain richer geometric information in rendering observations and semantic cues in RGB observations, leading to larger divergence in their representations across modalities.
When computing the local weight based on the holistic semantic consistency across modalities, these regions generate more dispersed yet salient responses, reflecting stronger feature representation discrepancies across modalities. 
As a result, the aggregated semantic consistency reflects stronger cross-modal variations, which leads to higher weights through the reweighting strategy.
This strategy emphasizes regions with pronounced cross-modal variations, encouraging the model to focus on informative and discriminative areas during the alignment stage.

With the incorporation of the reweighting strategy, the optimization process adaptively redistributes the contribution of each patch, thereby assigning greater importance to regions that are more challenging for cross-modal alignment.
As a result, these regions receive increased attention, leading to more effective alignment and improved learning of semantic cues from the RGB modality.
Therefore, by leveraging globally aggregated semantic consistency, the reweighting enhances the robustness of cross-modal alignment, enabling the model to better handle cross-modal discrepancies and ultimately improving the quality of the RGB-guided learned representations.

\section{Impact of the Number of Views}
We investigate the impact of the number of views on model performance, with the results shown in Figure \ref{fig:appendix_study_number_of_views}. 
As illustrated, increasing the number of views leads to consistent performance improvements across all metrics, 
as more views provide richer information for both rendering and RGB modalities, enabling the model to capture more detailed geometric and semantic cues. 
However, when the number of views increases from 9 to 11, a slight performance drop is observed. 
This demonstrates that incorporating excessive views may introduce redundant and noisy information, which can negatively affect model performance.

\section{Impact of Prompt Length}
\begin{figure}[t]
\centering
\begin{minipage}[t]{0.48\linewidth}
    \centering
    \includegraphics[width=\linewidth]{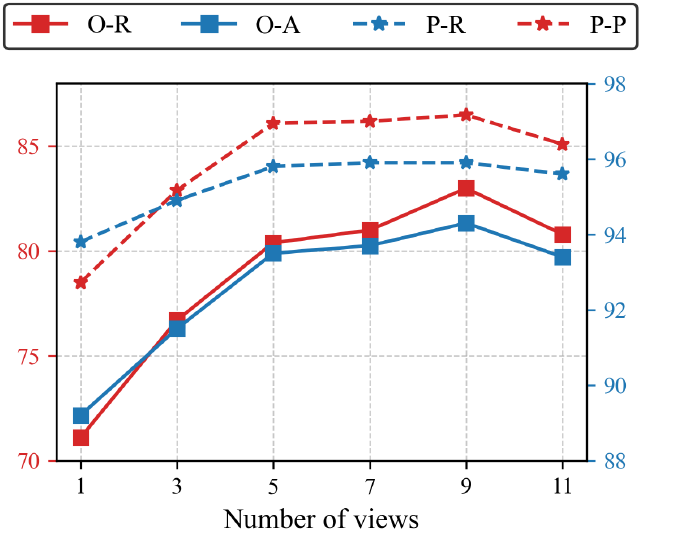}
    \caption{Impact of number of views on anomaly detection performance.}
    \label{fig:appendix_study_number_of_views}
\end{minipage}
\hfill
\begin{minipage}[t]{0.48\linewidth}
    \centering
    \includegraphics[width=\linewidth]{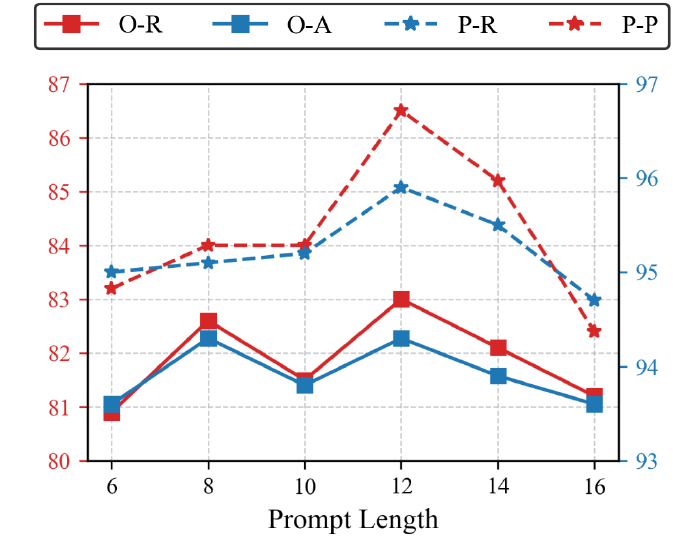}
    \caption{Impact of prompt length on anomaly detection performance.}
    \label{fig:appendix_study_length_of_prompt}
\end{minipage}
\end{figure}

We study the impact of prompt length on model performance, with the results shown in Figure \ref{fig:appendix_study_length_of_prompt}. 
As the length of learnable prompts increases, the model achieves improved performance, indicating that longer prompts help capture richer normal and anomalous semantics. 
However, when the length further increases (e.g., from 12 to 16), a performance drop is observed. 
This confirms that excessively long prompts may introduce redundant information, while overly short prompts are insufficient to model anomaly semantics. 
An appropriate prompt length (i.e., 12) achieves the best overall performance.

\section{Detailed Results}

For the one-vs-rest setting, we provide detailed per-category results on MVTec3D-AD in Table \ref{tab:appendix_One_vs_rest_category_results_MVTec} and Eyecandies in Table \ref{tab:appendix_One_vs_rest_category_results_Eyecandies}, including all evaluation metrics and their mean values to offer a more comprehensive analysis of the model performance.
For the cross-dataset setting, we also provide detailed per-category results on Eyecandies in Table \ref{tab:appendix_Cross_dataset_category_results_Eyecandies} and Real3D-AD in Table \ref{tab:appendix_Cross_dataset_category_results_real3d_ad}, including all evaluation metrics and their mean values.

\begin{table}[ht]
\centering
\caption{Per-category results on MVTec3D-AD under the one-vs-rest setting.}
\resizebox{\linewidth}{!}{
\begin{tabular}{cccccccccccc}
\hline
\textbf{Metric} & \textbf{Bagel} & \textbf{Cable gland} & \textbf{Carrot} & \textbf{Cookie} & \textbf{Dowel} & \textbf{Foam} & \textbf{Peach} & \textbf{Potato} & \textbf{Rope} & \textbf{Tire} & \textbf{Mean} \\ \hline
O-R             & 98.8           & 63.7                 & 96.5            & 94.4            & 73.2           & 61.3          & 90.9           & 99.0            & 92.6          & 59.8          & 83.0          \\
O-A             & 99.7           & 89.3                 & 99.3            & 98.5            & 92.2           & 86.7          & 97.4           & 99.8            & 97.1          & 83.2          & 94.3          \\
P-R             & 98.7           & 93.1                 & 99.4            & 89.0            & 95.8           & 89.7          & 99.2           & 99.8            & 99.4          & 95.3          & 95.9          \\
P-P             & 97.4           & 73.1                 & 97.8            & 79.4            & 85.6           & 63.7          & 97.8           & 99.3            & 94.0          & 77.3          & 86.5          \\ \hline
\end{tabular}
}
\label{tab:appendix_One_vs_rest_category_results_MVTec}
\end{table}

\begin{table}[ht]
\centering
\caption{Per-category results on Eyecandies under the one-vs-rest setting.}
\resizebox{\linewidth}{!}{
\begin{tabular}{cccccccccccc}
\hline
\textbf{Metric } & \textbf{Candy Cane} & \textbf{\begin{tabular}[c]{@{}c@{}}Chocolate\\ Cookie\end{tabular}} & \textbf{\begin{tabular}[c]{@{}c@{}}Chocolate\\ Praline\end{tabular}} & \textbf{Confetto} & \textbf{Gummy Bear} & \textbf{\begin{tabular}[c]{@{}c@{}}Hazelnut\\ Truffle\end{tabular}} & \textbf{\begin{tabular}[c]{@{}c@{}}Licorice\\  Sandwich\end{tabular}} & \textbf{Lollipop} & \textbf{Marshmallow} & \textbf{\begin{tabular}[c]{@{}c@{}}Peppermint\\ Candy\end{tabular}} & \textbf{Mean} \\ \hline
O-R             & 54.1                & 52.0                                                                & 71.3                                                                 & 78.1              & 72.0                & 66.2                                                                & 79.1                                                                  & 81.8              & 77.8                 & 80.4                                                                & 71.3          \\
O-A             & 53.9                & 60.2                                                                & 81.5                                                                 & 85.3              & 79.0                & 70.1                                                                & 86.0                                                                  & 78.9              & 84.3                 & 86.6                                                                & 76.6          \\
P-R             & 97.9                & 92.3                                                                & 90.0                                                                 & 94.2              & 90.2                & 87.9                                                                & 93.3                                                                  & 97.6              & 87.9                 & 95.1                                                                & 92.6          \\
P-P             & 87.1                & 72.9                                                                & 66.6                                                                 & 72.4              & 73.0                & 62.0                                                                & 74.1                                                                  & 84.1              & 64.6                 & 84.8                                                                & 74.2          \\ \hline
\end{tabular}
}
\label{tab:appendix_One_vs_rest_category_results_Eyecandies}
\end{table}

\begin{table}[ht]
\caption{Per-category results on Eyecandies under the cross-dataset setting.}
\resizebox{\linewidth}{!}{
\begin{tabular}{cccccccccccc}
\hline
\textbf{Metric} & \textbf{Candy Cane} & \textbf{\begin{tabular}[c]{@{}c@{}}Chocolate\\ Cookie\end{tabular}} & \textbf{\begin{tabular}[c]{@{}c@{}}Chocolate\\ Praline\end{tabular}} & \textbf{Confetto} & \textbf{Gummy Bear} & \textbf{\begin{tabular}[c]{@{}c@{}}Hazelnut\\ Truffle\end{tabular}} & \textbf{\begin{tabular}[c]{@{}c@{}}Licorice\\  Sandwich\end{tabular}} & \textbf{Lollipop} & \textbf{Marshmallow} & \textbf{\begin{tabular}[c]{@{}c@{}}Peppermint\\ Candy\end{tabular}} & \textbf{Mean} \\ \hline
O-R             & 53.2                & 49.1                                                                & 75.9                                                                 & 80.3              & 72.1                & 63.2                                                                & 79.4                                                                  & 75.6              & 80.3                 & 77.8                                                                & 70.7          \\
O-A             & 55.7                & 53.8                                                                & 84.3                                                                 & 86.4              & 78.6                & 68.9                                                                & 85.1                                                                  & 70.6              & 84.9                 & 84.4                                                                & 75.3          \\
P-R             & 97.8                & 91.5                                                                & 91.2                                                                 & 92.9              & 89.2                & 86.9                                                                & 93.3                                                                  & 98.0              & 87.4                 & 91.6                                                                & 92.0          \\
P-P             & 86.8                & 71.0                                                                & 70.0                                                                 & 69.0              & 69.5                & 61.5                                                                & 72.7                                                                  & 87.6              & 62.8                 & 74.8                                                                & 72.6          \\ \hline
\end{tabular}
}
\label{tab:appendix_Cross_dataset_category_results_Eyecandies}
\end{table}

\begin{table}[ht]
\caption{Per-category results on Real3D-AD under the cross-dataset setting.}
\resizebox{\linewidth}{!}{
\begin{tabular}{cccccccccccccc}
\hline
\textbf{Metric} & \textbf{Airplane} & \textbf{Car} & \textbf{Candybar} & \textbf{Chicken} & \textbf{Diamond} & \textbf{Duck} & \textbf{Fish} & \textbf{Gemstone} & \textbf{Seahorse} & \textbf{Shell} & \textbf{Starfish} & \textbf{Toffees} & \textbf{Mean} \\ \hline
O-R             & 50.7              & 71.4         & 86.5              & 60.5             & 100.0            & 47.4          & 82.5          & 84.6              & 79.5              & 87.0           & 84.2              & 78.5             & 76.1          \\
O-A             & 58.1              & 74.7         & 89.9              & 66.3             & 100.0            & 54.8          & 85.8          & 84.3              & 84.0              & 85.0           & 87.2              & 82.0             & 79.3          \\
P-R             & 63.2              & 69.9         & 83.4              & 76.0             & 93.9             & 49.5          & 82.6          & 84.8              & 80.7              & 78.3           & 80.0              & 77.5             & 76.7          \\ \hline
\end{tabular}
}
\label{tab:appendix_Cross_dataset_category_results_real3d_ad}
\end{table}

\section{Failure Cases}
\label{sec:appendix_failure_cases}
\begin{figure}[h]
\centering
\includegraphics[width=\linewidth]{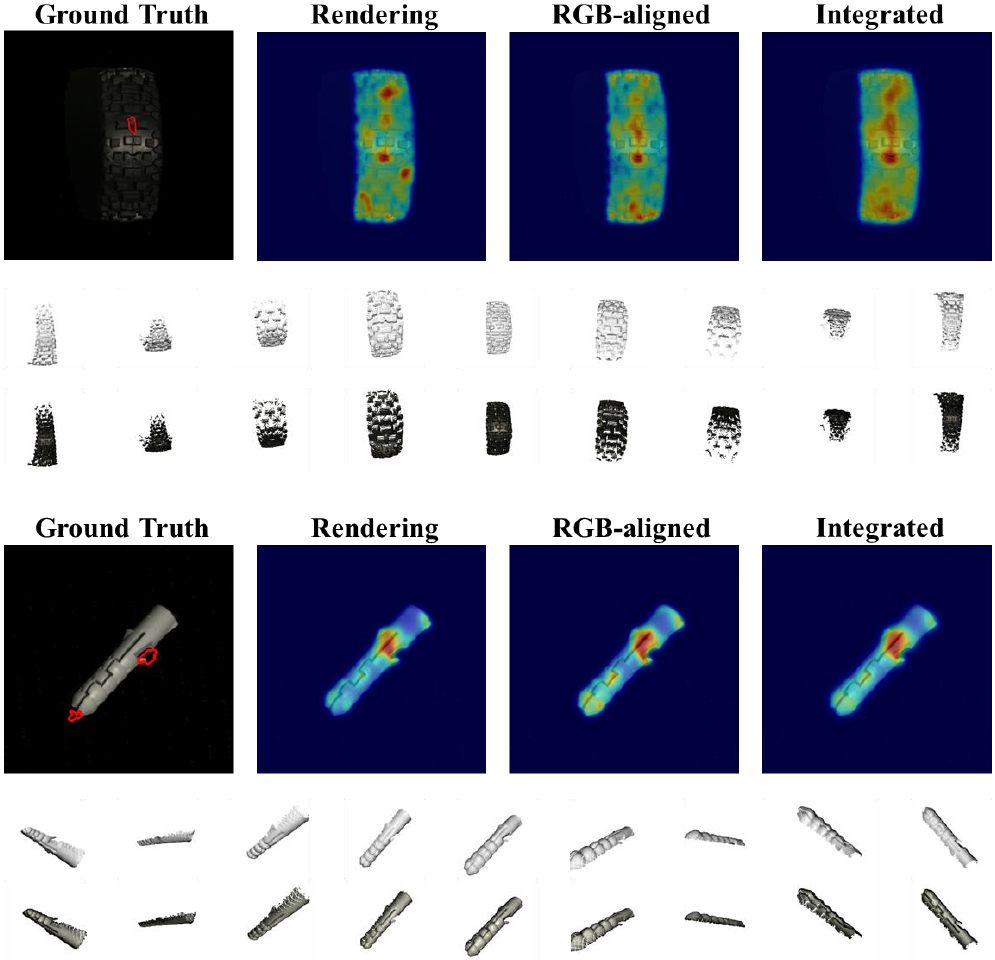}
\caption{Failure cases of Align3D-AD on MVTec3D-AD.}
\label{fig:Appendix_failure_cases}
\end{figure}

As illustrated in Figure \ref{fig:Appendix_failure_cases}, we observe that the characteristics of datasets can lead to incomplete or degraded rendering results in both RGB and rendering modalities. 
Since our framework relies on accurate cross-modal feature alignment, high-quality rendered observations are essential for learning consistent and accurate feature representations. 
Therefore, when the rendering quality is insufficient, these defects can adversely affect the model in the following aspects. First, the incomplete structures may introduce misleading and inaccurate signals, resulting in incorrect anomaly predictions. Second, the feature aligner fails to establish reliable correspondence between modalities, making it difficult to learn effective RGB-aligned features. 
As a result, the overall detection performance degrades on such challenging samples.

\section{Implementation Details for Complexity Analysis}
\label{sec:appendix_implementation_details_for_complexity_analysis}
We compare the computational efficiency of our method with PointAD \cite{PointAD} and GS-CLIP \cite{GS-CLIP} on MVTec3D-AD. 
Since PointAD and GS-CLIP report complexity analysis results on NVIDIA RTX 3090 GPUs in the original papers, we re-evaluate these methods on our hardware using the same 3D modality setting for a fair comparison. 
Specifically, all experiments are conducted on an NVIDIA RTX 4090 GPU, and the GPU is kept free during evaluation to avoid interference from other processes.
The batch size during inference is set to 1 for all methods.
The results reported in Table \ref{tab:ablation_study_complexity_analysis} reflect the efficiency of all methods under the same experimental conditions.
As shown in Table \ref{tab:ablation_study_complexity_analysis}, our method exhibits slightly higher computational costs compared with PointAD, which relies solely on rendering images. 
In contrast, when compared with GS-CLIP, which leverages multiple modalities including point clouds, rendering images, and depth maps, our method demonstrates consistently better efficiency. 
Generally, our approach achieves a more favorable balance between computational cost and accuracy, highlighting the practical effectiveness of our model.

\section{Additional Visualization Results}
\begin{figure}[h]
\centering
\includegraphics[width=\linewidth]{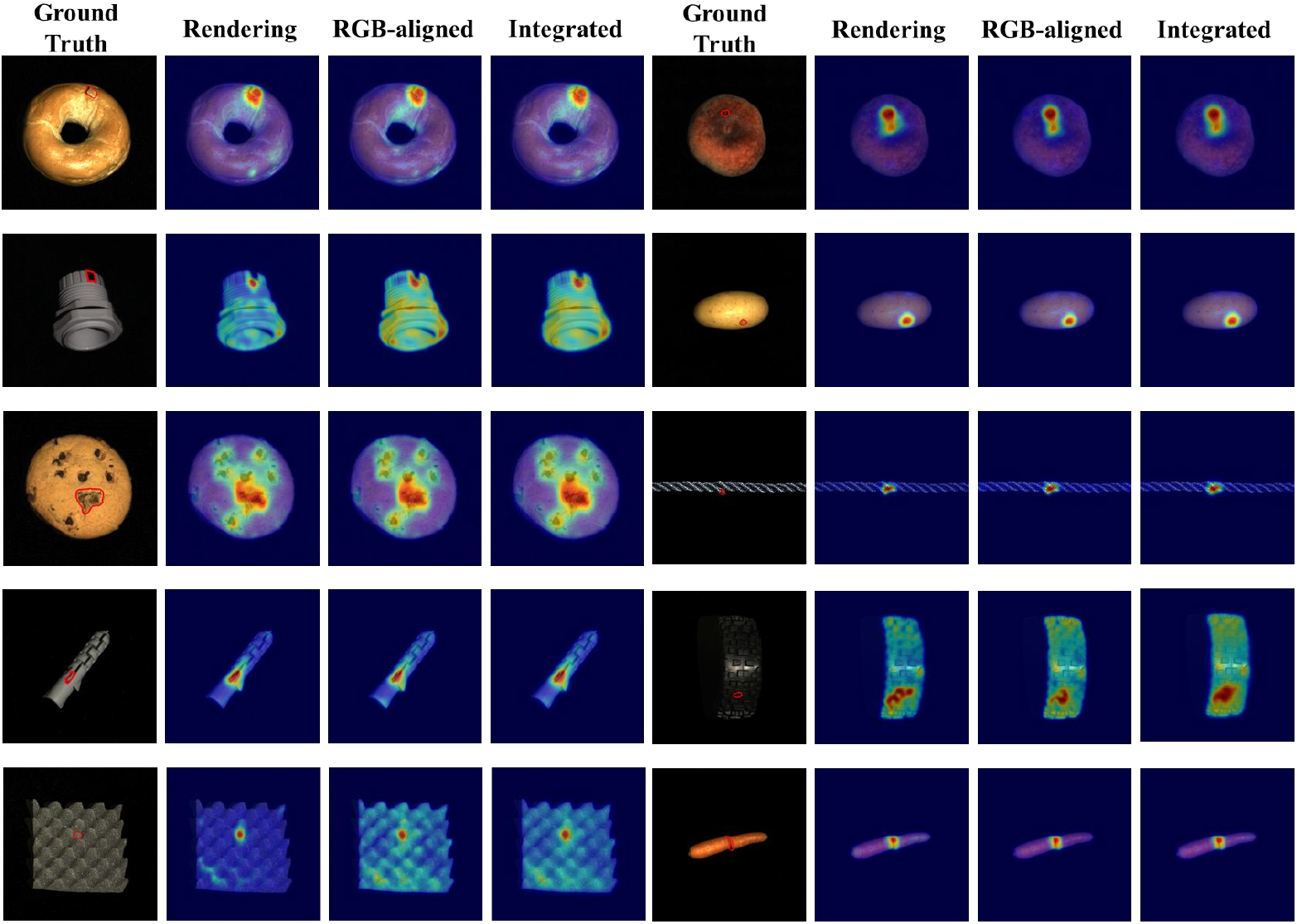}
\caption{Visualization of anomaly score maps on MVTec3D-AD.}
\label{fig:Appendix_MVTec_anpmaly_score_map}
\end{figure}

\begin{figure}[h]
\centering
\includegraphics[width=\linewidth]{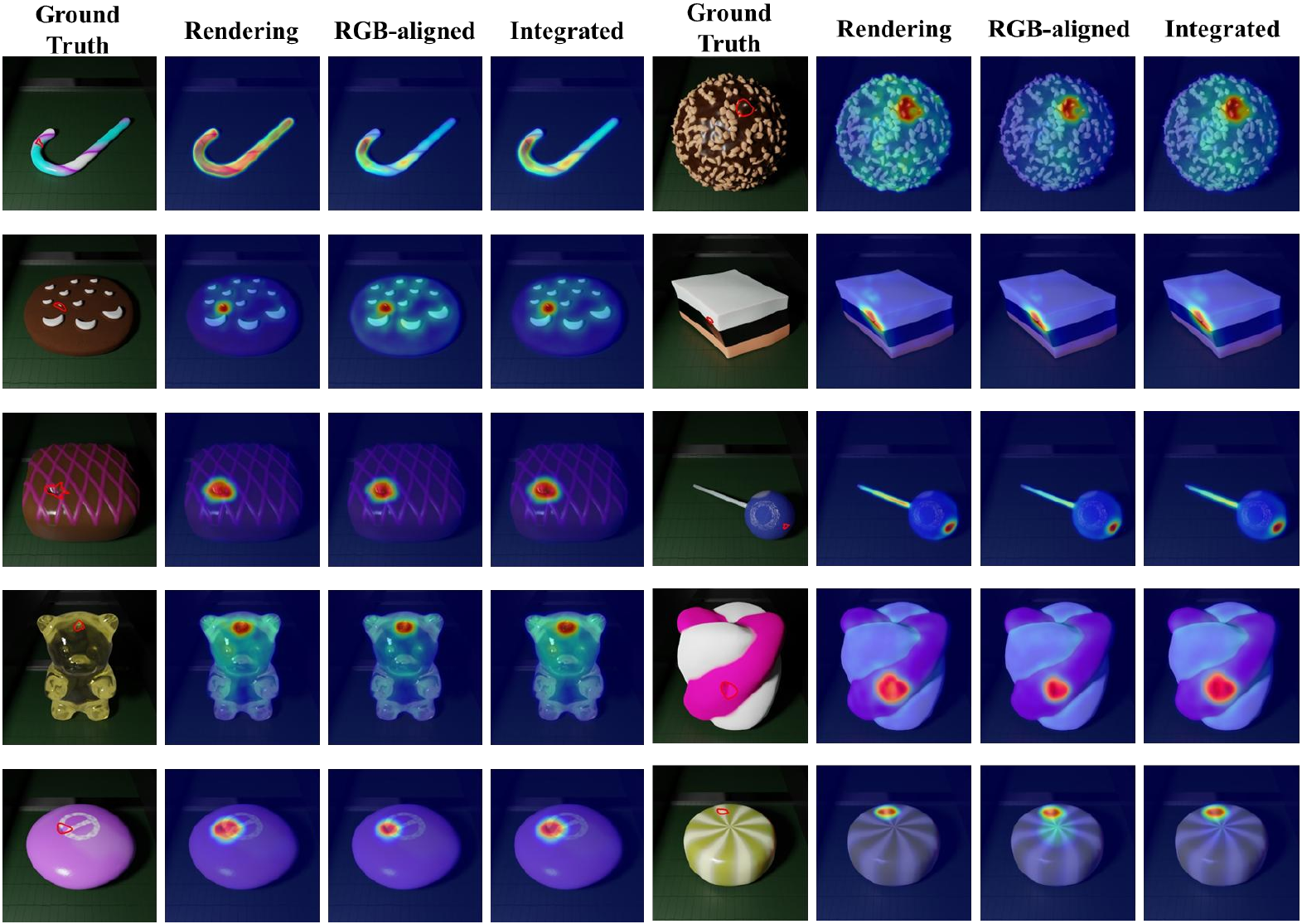}
\caption{Visualization of anomaly score maps on Eyecandies.}
\label{fig:Appendix_Eyecandies_anpmaly_score_map}
\end{figure}

We also provide more zero-shot visualization results of our model on MVTec3D-AD in Figure \ref{fig:Appendix_MVTec_anpmaly_score_map} and Eyecandies in Figure \ref{fig:Appendix_Eyecandies_anpmaly_score_map}. 
From the additional visualization results, we observe that both modalities are informative while exhibiting subtle differences across samples.
Although rendering representations effectively capture geometric and structural information, they are limited in expressing semantic cues, which may lead to suboptimal predictions in scenarios with ambiguous or subtle anomalies. 
In contrast, RGB-aligned representations provide richer semantic information, improving the recognition of semantically ambiguous defects that are difficult to distinguish from geometric cues alone. 
By integrating both representations, the model leverages the complementary information and achieves superior overall performance compared to the rendering-only feature representation, resulting in more reliable anomaly detection across all metrics.

\FloatBarrier



\end{document}